\documentclass[a4paper,fleqn]{cas-dc}
\usepackage[authoryear,longnamesfirst]{natbib}
\usepackage{amsmath,amsfonts}
\usepackage{algorithmic}
\usepackage{algorithm}
\usepackage{array}
\usepackage[caption=false,font=normalsize,labelfont=sf,textfont=sf]{subfig}
\usepackage{textcomp}
\usepackage{stfloats}
\usepackage{url}
\usepackage{graphicx}
\usepackage{makecell}
\usepackage{multirow}
\usepackage{booktabs}
\graphicspath{{figures/}}
\usepackage{color}
\usepackage{hyperref}
\usepackage{float}
\usepackage{verbatim} 
\usepackage{apalike}

\begin{document}
\let\WriteBookmarks\relax
\def\floatpagepagefraction{1}
\def\textpagefraction{.001}


\shorttitle{Self-Supervised Learning for Point Clouds Data: A Survey}
\shortauthors{Changyu Zeng et~al.}

\title[mode = title]{Self-Supervised Learning for Point Clouds Data: A Survey}

\author[1,2,3]{Changyu Zeng}

\author[2]{Wei Wang}

\author[3]{Anh Nguyen}

\author[1,2,3]{Yutao Yue}[orcid=0000-0003-4532-0924]
\ead{yueyutao@idpt.org}
\cormark[1]

\affiliation[1]{organization={Institute of Deep Perception Technology, JITRI},
    city={Wuxi},
    postcode={214000},
    country={China}}

\affiliation[2]{organization={Department of Computing, School of Advanced Technology, Xi'an Jiaotong-Liverpool University},
    city={Suzhou},
    postcode={215123},
    country={China}}

\affiliation[3]{organization={Department of Computer Science, University of Liverpool},
    city={Liverpool},
    postcode={L69 7ZX},
    country={United Kingdom}}

\cortext[cor1]{Corresponding author}

\begin{abstract}
    3D point clouds are a crucial type of data collected by LiDAR sensors and widely used in transportation applications due to its concise descriptions and accurate localization. Deep neural networks (DNNs) have achieved remarkable success in processing large amount of disordered and sparse 3D point clouds, especially in various computer vision tasks, such as pedestrian detection and vehicle recognition. Among all the learning paradigms, Self-Supervised Learning (SSL), an unsupervised training paradigm that mines effective information from the data itself, is considered as an essential solution to solve the time-consuming and labor-intensive data labelling problems via smart pre-training task design. This paper provides a comprehensive survey of recent advances on SSL for point clouds. We first present an innovative taxonomy, categorizing the existing SSL methods into four broad categories based on the pretexts' characteristics. Under each category, we then further categorize the methods into more fine-grained groups and summarize the strength and limitations of the representative methods. We also compare the performance of the notable SSL methods in literature on multiple downstream tasks on benchmark datasets both quantitatively and qualitatively. Finally, we propose a number of future research directions based on the identified limitations of existing SSL research on point clouds.
\end{abstract}


\maketitle

\section{Introduction}
With the rapid development of 3D data processing technologies, an increasing number of relevant applications have emerged in both industrial and daily usage, such as indoor navigation \citep{el2021indoor}, autonomous driving \citep{YingLi2020DeepLF}, and object modeling \citep{yang2019reppoints}. LiDAR is one of the indispensable types of sensors to capture disordered 3D point cloud data from traffic scenes, which has enabled more challenging tasks like pedestrian detection \citep{8078512} and road semantic segmentation \citep{wu2019squeezesegv2} based on the strong inference ability of deep neural networks (DNNs).

\begin{figure}
    \centering
    \includegraphics[width=0.97\linewidth]{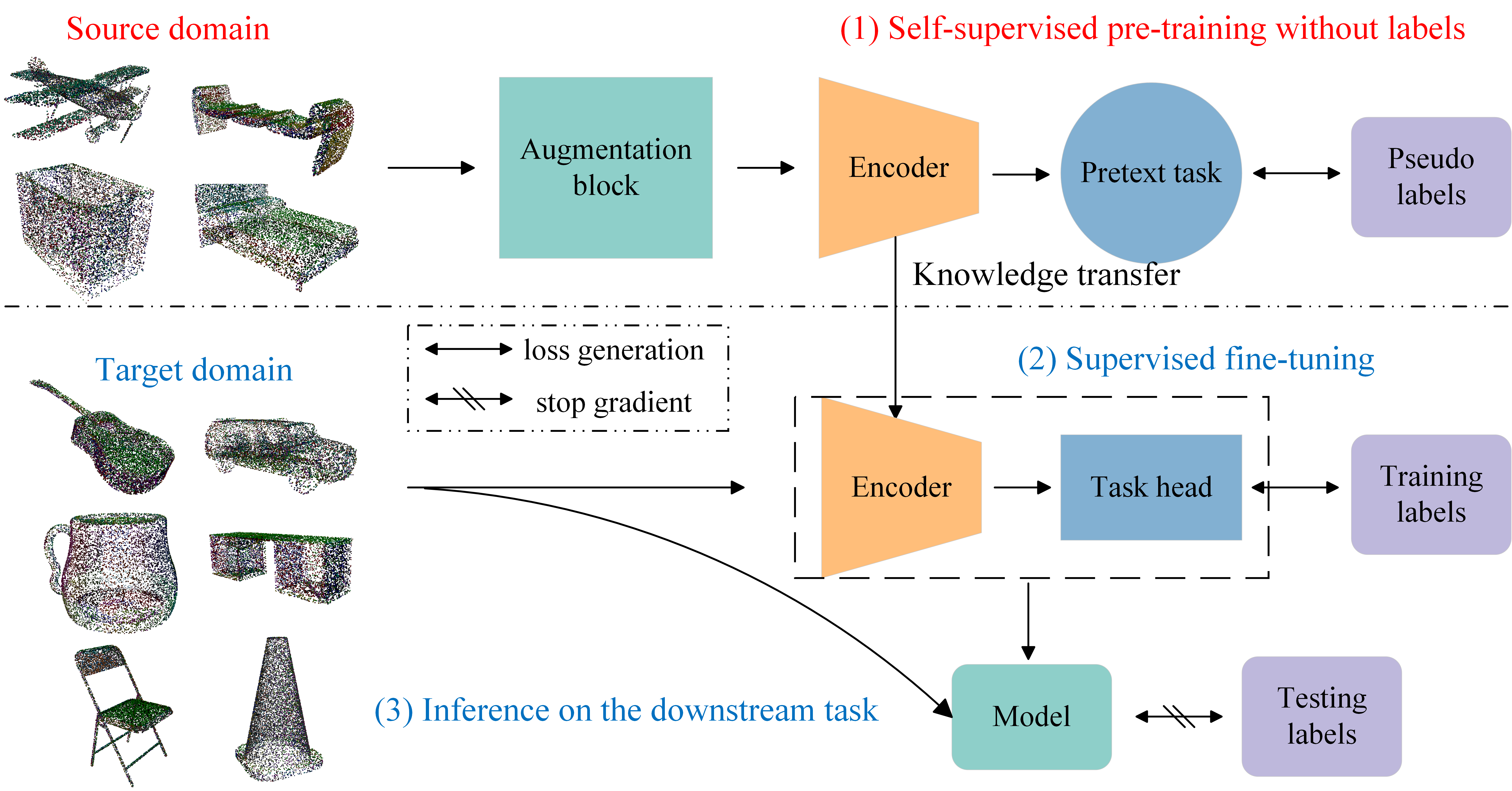}
    \caption{The general pipeline of SSL used in the point cloud data. (1) Pre-training stage: point cloud data is firstly pre-processed through the augmentation block and then fed into the point-specific encoder to learn feature representations. The features are utilized to complete well-design pretext tasks, where the output will be compared with the pseudo labels derived from the original data to generate a loss and to update encoder parameters via back-propagation; (2) Supervised fine-tuning stage: the well-trained encoder is transferred to the target domain. A task head is trained with the training labels in a supervised manner to complete the downstream tasks; (3) Inference stage: the encoder and task head are concatenated as a model to execute inference on the test set. The effectiveness of the SSL pre-training framework can be evaluated based on the performance of the model on the downstream tasks.}
    \label{fig:pipeline}
\end{figure}

However, several well known problems in the supervised point cloud DNNs hinder their further development and practical uses. For example, accurate environment perception via DNNs requires millions of labeled data as the input, while point cloud annotating is labor-intensive and time-consuming due to its disordered and sparse nature \citep{dai2017scannet}. Besides, manual labeling by human experts or users inevitably leads to mistakes such as mislabeling and omission. Another long-standing problem is that the supervised learning paradigm struggles to capture the underlying patterns of new data and fails to generalize the pre-training model to downstream tasks because of overfitting caused by noisy labels \citep{sariyildiz2022improving}.

The aforementioned issues motivate research in extracting effective feature representations from point clouds via Self-Supervised Learning (SSL) to learn implicit while better representations without manual annotations. Not only does it solve the problem of the error-prone and expensive labeling process, but also relieve the domain adaptation (DA) issues \citep{csurka2017domain} with improved model generalization ability. Under the SSL paradigm, basic geometric as well as advanced semantic information can be extracted as knowledge and migrated to downstream tasks under the transfer learning setup. This process approximates human learning that discovers objective principles of the world by observing phenomena and summarizing them into a system of experience and knowledge.

Fig. \ref{fig:pipeline} shows a general pipeline of SSL on point cloud data. The goal of SSL is to pre-train an encoder on an unlabeled, large-scale point cloud dataset (source domain), and to transfer the well-trained network to other datasets (target domain) in various downstream tasks. A complete SSL framework usually contains the following important modules.

\begin{itemize}
    \item \textbf{Data augmentation}: The raw input is augmented via some easy-to-implement pre-processing operations such as translation, rotation, flip, and adding noise \citep{zhang2022point}. The objective is to expand the size and diversity of the raw data and to provide subjects for subsequent pretext tasks. The details will be discussed in Section \ref{sec:augmentation}. 
    \item \textbf{Encoder}: The encoder is a point-specific deep network that captures the hierarchical representation of the input point cloud data. We will introduce some commonly used point cloud encoders that learn either from downsampling layer-by-layer \citep{qi2017pointnet,qi2017pointnet++} or  local areas to capture the association between different blocks \citep{YinZhou2018VoxelNetEL,wang2019dynamic}. The details will be discussed in Section \ref{sec:models}. 
    \item \textbf{Pretext task}: At the core of the framework is the design of a pretext task that mines the hidden self-supervision signal via the interactions between the encoder and data. This part is also the focus of the survey and will be discussed in detail in Section \ref{sec:pretext}.
    \item \textbf{Knowledge transfer}: The well-trained encoder will be transferred to another dataset with the knowledge gained in the source domain after completing the pretext task. A task head is constructed and trained by a small amount of labelled data in the target domain as the supervision signals to fine-tune the whole architecture. The details will be discussed in Section \ref{sec:downstream}.
    \item \textbf{Downstream task}: To evaluate the effectiveness of the SSL framework, the pre-trained encoder will be transferred and evaluated on another dataset for performance evaluation, e.g. object classification, part segmentation, and object detection. The details will be discussed in Section \ref{sec:downstream}.
\end{itemize}

\begin{figure}
    \centering
    \includegraphics[width=0.97\linewidth]{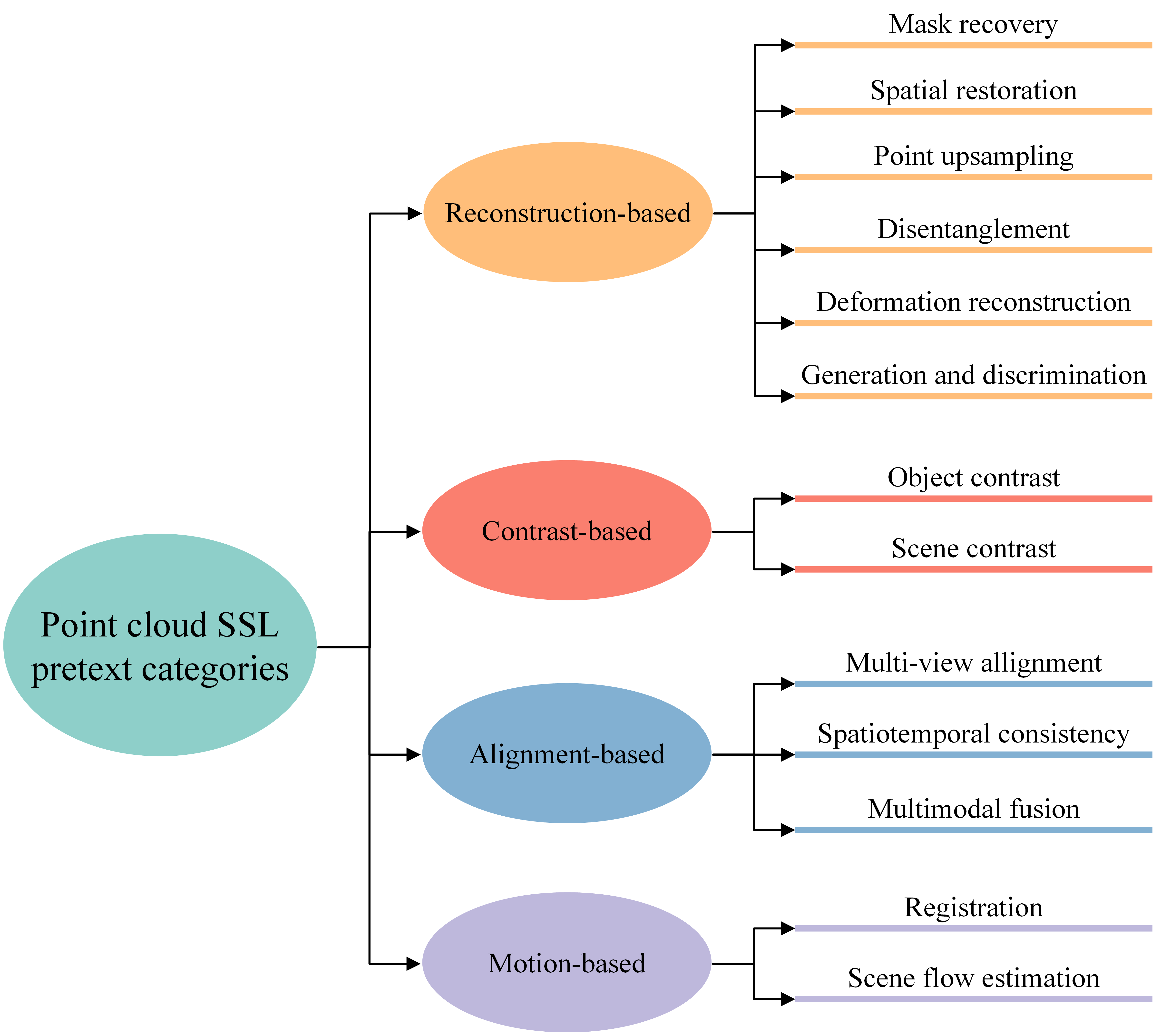}
    \caption{Taxonomy of SSL for point cloud data based on pretext tasks.}
    \label{fig:taxonomy}
\end{figure}

Thriving progress has been made on point cloud SSL recently, and new models, algorithms and benchmark datasets are emerging quickly and continuously. A systematic review on this exciting topic, especially the research published in the past three years, is urgently needed. In our study, we find that the survey in \citep{xiao2022unsupervised} employed a similar methodology but focused on unsupervised representation learning. However, it lacked a review on the state-of-the-art SSL models, and in particular, a detailed demonstration of most recent published works. Therefore, we are motivated to provide a comprehensive review on the recently published, representative research on point cloud SSL. Our contributions can be summarized as follows:

\begin{itemize}
    \item \textbf{Systematic and novel taxonomy:}
    We propose a novel and systematic taxonomy for categorizing the diverse kinds of point cloud SSL methods to provide a clear and holistic view on the state of the art. Taking into consideration the characteristics of popular pretext tasks, the taxonomy groups current methods into four broad categories. Each broad category is further subdivided into more fine-grained sub-categories according to the methods in feature utilization as shown in Fig. \ref{fig:taxonomy}.
    \item \textbf{Comprehensive and detailed summary:}
    We conduct a comprehensive review of the state of the art, including the background of SSL and point clouds, commonly used datasets and models, pretext tasks, and downstream tasks with performance comparison.
    \item \textbf{Exhaustive dataset summary and evaluation comparison:}
    We summarize the unique characteristics of 18 most frequently utilized datasets in the point cloud research. More importantly, we compare the performance of different SSL methods on these datasets according to various downstream tasks.
    \item \textbf{Future directions:}
    Based on our investigation, we summarize and discuss the major limitations and challenges in the current research and propose potential future directions which would hopefully motivate more theoretical and practical research towards more intelligent and effective SSL approaches for point cloud data processing.
\end{itemize}

The rest of the paper is organized as follows: Section \ref{sec:background} introduces the preliminaries for this survey to equip the readers with the necessary background knowledge on SSL and point cloud data. Section \ref{sec:pretext} represents the main body of the study and provides an exhaustive and detailed analysis on the state of the arts methods according to the structure of the proposed taxonomy. Section \ref{sec:downstream} illustrates an evaluation and comparison study on the performance of different SSL methods on the frequently utilized downstream tasks and benchmark datasets. Section \ref{sec:future} discusses the limitations and challenges of current research and proposes potential future directions, and Section \ref{sec:conclusion} concludes the paper.

\section{Background} \label{sec:background}

\subsection{Self-supervised learning in the language and image domain}
We firstly describe the development history of SSL in the language and image domains. The purpose is to provide readers a general understanding on SSL. Although data types vary from domains to domains, the core idea of SSL remains the same: to leverage data characteristics for transformation processing and to make the transformed data consistent with the original input in terms of feature representation by contrasting or reconstruction.

The idea of SSL was firstly introduced in Natural Language Processing (NLP) research. After converting words into vectors, e.g. Word2Vec \citep{mikolov2013efficient}, and utilizing the relationships between the representations and context, models could learn semantic representations from neighboring words or sentences through pretext task formulations such as next sentence prediction \citep{devlin2018bert}, auto-regressive language modeling \citep{floridi2020gpt}, or sentence permutation \citep{lewis2019bart}. Landmark models such as GPT \citep{floridi2020gpt} and BERT \citep{devlin2018bert}, and many variants celebrate great achievements in not only NLP but also other fields later.

In the field of image processing and computer vision, different SSL methods impose simple variations on image data and extract features by recovering it to the original input, for example, from simple tasks like relative position prediction \citep{doersch2015unsupervised,noroozi2016unsupervised} and rotation angle prediction \citep{SpyrosGidaris2018UnsupervisedRL}, to reconstructing blocks masked by surrounding visible pictures \citep{DeepakPathak2016ContextEF,he2021masked}. Free semantic label-based \citep{faktor2014video,stretcu2015multiple,croitoru2017unsupervised,jiang2018self} and cross-modal-based methods \citep{arandjelovic2017look,agrawal2015learning,jayaraman2015learning} have been proposed, which learn representations via automatically generated semantic labels and extra information from other modalities. Recently, the research community shows a great interest on contrastive learning \citep{chen2020simple,he2020momentum,caron2020unsupervised}, which aims to differentiate positive and negative samples by comparison using data augmentation techniques. These research works inspired the study of SSL on point clouds, with similar ideas transferred from 2D to 3D by adapting for data peculiarities.

\subsection{Properties of the point cloud data} \label{sec:properties}

         


Data properties are distinct between naturals languages, images, and point clouds. Languages are usually complex and abstract in nature, and contain ambiguous information due to its versatility and richness. It is expressed in a sequence of words, which is discrete and unstructured in the representation space \citep{he2020momentum}. In contrast, images contain rich visual information, such as color, texture, and shape information of an object in high-dimensional space \citep{jing2020self} for human perception. They are usually represented as 2D data by using a matrix of pixel values.

Simply speaking, point cloud data is similar to image data in terms of visual format and can be regarded as 3D stereo images with depth information. However, the attributes of point cloud data are completely different in geometric representation. Specifically, a point cloud is a collection of discrete, disordered, and topology-free 3D points. The most basic information contained in the points is the position coordinates $(x_i, y_i, z_i)$ in the Euclidean space, where $i$ is the number of points in the object. There are also other optional attributes such as color, intensity, reflectivity, etc., specifying physical properties of the points in more detail. The input order is trivial for point cloud data and does not impact the semantic meaning while it is crucial for images and language where various words or pixel sequences lead to completely divergent connotations. Additionally, point cloud data is invariant to rigid transformation, which means that it remains unchanged after rotation and translation. Some of such exclusive properties can be summarized as follows:

\begin{itemize}
    \item \textbf{Sparsity:} The point cloud data is discretely distributed on the surface of the scanned object or scene.
    \item \textbf{Non-uniformity:} The distance between points is not fixed and is determined by various factors such as the instruments' sampling strategy, relative position, and scanning range.
    \item \textbf{Imcomplete data:} Some parts of real-scanned surfaces are incomplete due to self or external occlusion.
    \item \textbf{Noise:} It is inevitable that noise from environmental factors or inaccuracies in instruments will be present.
    \item \textbf{Permutation invariance:} The order of points does not affect the overall semantic representation of point cloud objects, so identical point cloud objects can be expressed by various matrices.
    \item \textbf{Transformation immutability:} Point clouds remain immutable through rigid transformations such as rotation and translation.
    \item \textbf{Points interaction:} There are correlations, either strong or weak, between points in global and local regions.
\end{itemize}

\subsection{Point Cloud Dataset}\label{sec:data}
Quality benchmark datasets (e.g. complete, well-varied, and densely-labeled) play essential roles in SSL research. This section lists the most commonly used point cloud datasets and summarize them in Table \ref{tab:dataset} in terms of sample number, object categories, suitable tasks, and highlights. These datasets may contain synthetic and real scanned data, in single frames and time series and from individual objects and complex scenes. There are also a few datasets for complex traffic scenarios (e.g. automatic driving) containing extra data in different modalities, such as from images or radars.

\begin{table*}[]
    \caption{Summary of commonly used point cloud datasets. Abbreviations for suitable tasks: Cls (Classification); Seg (Semantic Segmentation); Det (Object Detection); Com (Semantic Scene Completion); Rec (Surface Reconstruction); CM (Cross-Modal tasks); Pos (Pose estimation); Tra (Object Tracking)}
    \label{tab:dataset}
    \centering
    \resizebox*{\linewidth}{!}{
    \begin{tabular}{|c|c|c|c|c|c|c|}\hline
        Year & Name & \#Samples & \#Categories & Types & Suitable tasks & Highlights\\\hline
        2012 & KITTI \citep{geiger2012we} & Over 200K objects & 8 & RGB & Cls/Det/CM & Comprehensive outdoor driving dataset \\ 
        2015 & ModelNet \citep{wu20153d} & 12,311 models & 40 & CAD & Cls/Seg/Rec & Frequently used in classification and few-shot \\
        2015 & ShapeNet \citep{chang2015shapenet} & 57,448 models & 55 & CAD & Cls/Seg & Commonly employed as the pre-training dataset \\
        2015 & SUN RGBD \citep{song2015sun} & 10,335 images & 37 & RGB-D & Seg/Det/Pos & A RGB-D scene understanding benchmark suite \\
        2016 & SceneNN \citep{hua2016scenenn} & 100 scenes & - & RGB-D & Det/Rec/Pos & Using unique triangle meshes shape contour\\
        2016 & ObjectNet3D \citep{xiang2016objectnet3d} & 44,147 shapes & 100 & CAD & Det/CM/Pos & Well-aligned 2D-3D dataset \\
        2016 & S3DIS \citep{armeni20163d} & 272 scans & 13 & Point & Cls/Seg/CM & Large-scale indoor space scanning dataset \\
        2017 & ScanNet \citep{dai2017scannet} & 1513 scenes & 20 & RGB-D & Cls/Seg/Com & Rich labels for scene understanding tasks \\
        2017 & Semantic3D.net \citep{hackel2017semantic3d} & Over 4B points & 8 & Point & Cls/Seg/Det & High quality resolution and scope outdoor dataset \\
        2018 & Pix3d \citep{sun2018pix3d} & 10,069 3D-2D pairs & 9 & CAD & Rec/CM/Pos & Pixel-level image-shape pairs dataset \\
        2019 & ABC \citep{koch2019abc} & 1M objects & - & CAD & Seg/Rec & Providing a benchmark for surface normal estimation \\
        2019 & ScanObjectNN \citep{uy2019revisiting} & 2,902 objects & 15 & Point & Cls/Seg & Challenging real-world scenario with noise \\ 
        2019 & PartNet \citep{mo2019partnet} & 26,671 models & 24 & Points & Seg/Rec & Producing fine-grained multi-level 3D part objects \\
        2020 & RobustPointSet \citep{taghanaki2020robustpointset} & 73,843 & 40 & Mesh & Cls & Benchmark to evaluate the robustness of classifiers \\
        2020 & Waymo \citep{sun2020scalability} & 12M objects & - & Point & Det/CM/Tra & Suitable for cross-modal and transfer learning \\
        2020 & NuScenes \citep{caesar2020nuscenes} & 1K scenes & 23 & Point & Det/CM/Tra & Containing additional annotations and scenes \\
        2021 & SensatUrban \citep{hu2021towards} & 4B points & 13 & Point & Seg & Data collected by UAV over UK landscape \\ 
        2022 & STPLS3D \citep{chen2022stpls3d} & 62 scenes & 18 & Point & Seg/Det & Covering both real and synthetic aerial point clouds \\
        \hline
        
    \end{tabular}}

\end{table*}

\begin{itemize}
    \item \textbf{KITTI} \citep{geiger2012we} is a benchmark suite for autonomous driving vision tasks. The dataset was collected using several pieces of equipment, including four video cameras, a laser scanner, and a localization system. It includes not only point clouds but also stereo and optical flow data. There are more than 200,000 annotated point cloud scenarios consisting of cars and pedestrians, providing a novel and challenging benchmark for 3D object detection and orientation estimation.
    \item \textbf{ModelNet} \citep{wu20153d} is the most widely used 3D point cloud CAD dataset for object classification and few-shot learning. It contains 12,311 single objects from 40 categories, with each point composed of six dimensions of information, including XYZ spatial coordinates and RGB values.
    \item \textbf{ShapeNet} \citep{chang2015shapenet} is a relatively large-scale repository of 3D CAD objects frequently employed as a pre-training dataset. It contains more than 3 million samples categorized into 55 classes under the WordNet synsets \citep{miller1995wordnet} criteria. The annotations in the dataset are versatile, including rigid alignments, parts, physical sizes, and key points.
    \item \textbf{SUN RGBD} \citep{song2015sun} is an RGB-D scene understanding benchmark suite containing 10,335 samples at a comparable scale to PASCAL VOC \citep{pascal-voc-2012}. It has 146,617 2D polygons and 64,595 3D bounding boxes densely annotated to indicate object orientation, room layout, as well as scene category for overall scene awareness.
    \item \textbf{S3DIS} \citep{armeni20163d} is a 3D indoor venue dataset that consists of scanning of 272 rooms in 6 areas overlaying a 6,000 $m^2$ area. It has 13 semantic categories labeled by fine-grained point-wise annotations carrying full 9D information, including XYZ, RGBs, and normalized location coordinates.
    \item \textbf{ScanNet} \citep{dai2017scannet} is a 3D RGB-D dataset that comprises 2.5M views in 1,513 scenes acquired in 707 indoor environments. Various tests containing semantic voxel labeling and CAD model retrieval proved that ScanNet provides quality data for 3D scene understanding.
    \item \textbf{ScanObjectNN} \citep{uy2019revisiting} was proposed as a collection of real-world indoor point cloud scenes to break the performance saturation of 3D object classification on synthetic data. This dataset introduces new challenges for 3D object classification due to the presence of background noise and occlusions that require networks' ability on context-based reconstructions and partial observations.
    \item \textbf{Waymo} \citep{sun2020scalability} is a large autonomous driving dataset produced by Waymo in collaboration with Google Inc. The dataset consists of 1,150 urban and suburban geography scenes spanning 20 seconds, which are collected via well-synchronized and calibrated LiDARs and cameras.
    \item \textbf{NuScenes} \citep{caesar2020nuscenes} is another remarkable multimodal dataset provided by the full sensor suite including cameras, radars, and LiDARs. Compared to other autonomous driving datasets, it contains additional annotations like pedestrian pose, vehicle state, and also scenes from nighttime and rainy weather.
\end{itemize}

\subsection{Point cloud data augmentation}\label{sec:augmentation}

\begin{figure}
    \centering
    \includegraphics[width=0.97\linewidth]{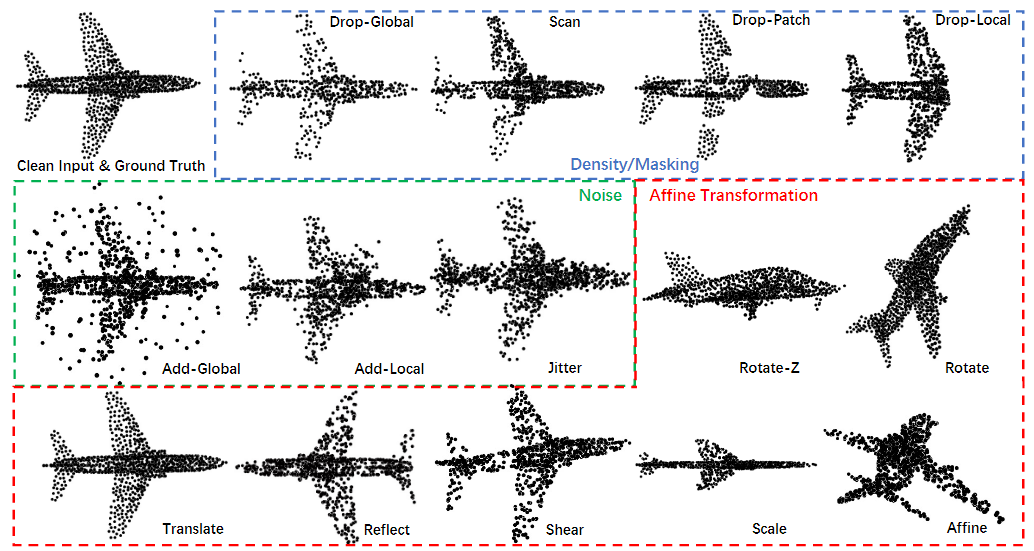}
    \caption{Illustration of the commonly used data augmentation methods for point cloud data. There are a total of 14 sub-categories of data augmentation methods that could be classified as three general corruption families. The figure is adapted from \citep{zhang2022point}.}
    \label{fig:augmentation}
\end{figure}

Data augmentation is a crucial technique for enhancing DNNs performance by increasing the amount and diversity of training samples. For SSL tasks, it not only prevents the model from overfitting but also facilitates capturing robust and invariant representations of point clouds under multiple transformations. In this section, we will introduce the commonly used data augmentation methods and compare the effectiveness of each methods via a metric called task relatedness.

Essentially, data augmentation is a process of generating new data by adding interventions or corruptions without destroying the original semantic expressions. For point clouds, augmentation methods are based on the properties mentioned in Section \ref{sec:properties} and can be classified into three general groups: density/masking, noise, and affine transformation \citep{zhang2022point}. These three corruption families could be further divided into 14 sub-categories as shown in Fig. \ref{fig:augmentation}.

Density/masking is the most frequent data augmentation method adopted in mask autoencoder (MAE) type SSL research \citep{he2021masked,pang2022masked,yu2021point}. Based on the principle that point cloud data is sparse with uneven density, randomly removing a certain percentage of points while preserving part of the semantic expression presents a challenging learning objective for such MAE-based tasks. On the contrary, the noise based methods impose interventions on the original clean input to increase the difficulty of feature extraction. Affine transformation leverages point cloud invariance characteristics to shift the spatial coordinates of each points. This has significant impact on the input since the basic position information completely changes.

The work in \citep{chen2020simple,zhang2022point} investigated the effectiveness of the aforementioned augmentation methods as pretext data preprocessing on downstream classification tasks. Task relatedness is employed as the evaluation metric to statistically measure the performance of SSL models on downstream tasks, which provides valuable advice for proxy data augmentation selection. Following \citep{zamir2018taskonomy}, for each pretext task $c$, its task relatedness to downstream task $t$ is defined as:

\begin{equation}
    A_{c \to t} := \mathbb{E}_{x \in X} \mathcal{I}_{t}(R_c(E_c(x)),f_t(x))
\end{equation}

Where $x$ is a sample in a point cloud dataset $X$; $E_c$ is the model's encoder pre-trained on task $c$; $R_c$ is a readout function, which indicates the classification head composed of several fully connected (FC) layers; $f_t$ is the labeling function; $\mathcal{I}_{t}$ is accuracy measurement estimating whether the downstream output $R_c(E_c(x))$ conforms to the ground truth $f_t(x)$.

\begin{figure}
    \centering
    \includegraphics[width=0.97\linewidth]{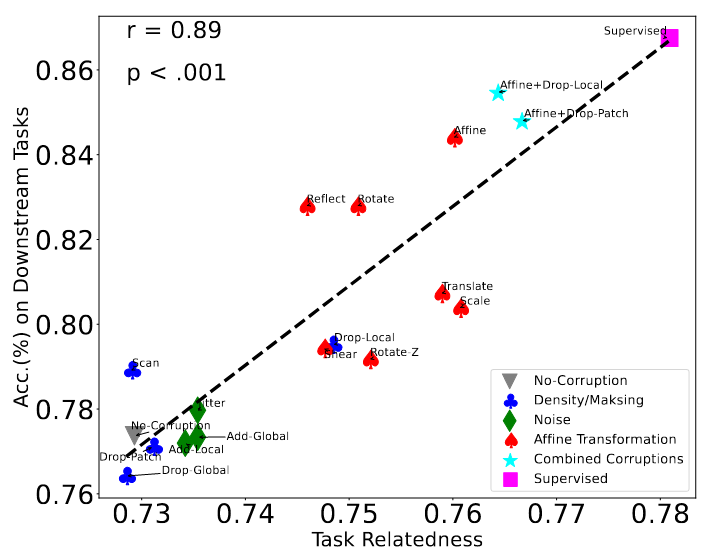}
    \caption{Illustration of the relationship between task relatedness and classification accuracy on downstream tasks. $r$ and $p$ are the coefficients to measure the linear relationship and statistical significance for the Pearson correlation, respectively. The figure is adapted from \citep{zhang2022point}.}
    \label{fig:task_relatedness}
\end{figure}

To further explore the relationship between task relatedness and classification accuracy on downstream tasks, Pearson correlation coefficient $r$ and $p$-value are utilized to estimate the linear relationship as well as statistical significance \citep{fraser1976probability}, respectively, where $|r| > 0.5$ refers to a strong correlation and $p < 0.05$ is considered statistically significant. Fig. \ref{fig:task_relatedness} demonstrates the statistically significant linear relationship between task relatedness and classification accuracy on downstream tasks when $r = 0.89$ and $p < 0.001$. The results reveal a counter-intuitive fact that frequently used density/mask and noise-based data augmentation methods are ineffective for downstream tasks either in accuracy and task relatedness. Conversely, the seemingly simple affine transformation enhances task relatedness to point cloud classification, resulting in higher accuracy. Furthermore, combining corruptions of affine transformation and mask can approach the performance of supervised benchmarks. Hence, using affine transformation-based methods for data augmentation is preferable for in SSL pre-training.

\subsection{Popular deep models for point clouds}\label{sec:models}
SSL techniques designed for languages and images need to be revised and extended for point clouds. For instance, traditional CNN networks cannot handle irregular and discrete point cloud data well since there is no guarantee that a corresponding point exists at the same relative position of the convolution. In this section, we briefly introduce five point cloud networks that are frequently used as feature extraction encoders in the literature and summarize their respective characteristics in Table \ref{tab:models}.

\begin{table*}[]
    \centering
    \caption{Commonly used deep networks for extracting point cloud features.}
    \label{tab:models}
    \resizebox*{\linewidth}{!}{
    \begin{tabular}{|c|c|c|c|} \hline
        Model & Year & Architecture & Contributions \\\hline
        PointNet \citep{qi2017pointnet} & 2017 & CNN & Pioneer in direct processing of raw point clouds with lightweight architecture \\
        PointNet++ \citep{qi2017pointnet++} & 2017 & CNN & Aggregating local neighborhood by multi-scale and multi-resolution sampling and groping \\ 
        VoxelNet \citep{YinZhou2018VoxelNetEL} & 2018 & 3D CNN & Partitioning disordered point clouds into regular voxels for local feature learning \\
        DGCNN \citep{wang2019dynamic} & 2019 & Graph CNN & Constructing a dynamic local graph to capture edge features around a neighbor \\
        PCT \citep{guo2021pct} & 2021 & Transformer & Successfully capturing the long-range dependencies between point patches \\
        GANs \citep{goodfellow2014generative} & 2014 & GAN & Generating synthetic data through adversarial training \\\hline
    \end{tabular}}

\end{table*}
    
\subsubsection{PointNet}
To reduce data size and computation complexity, Qi et al. proposed PointNet \citep{qi2017pointnet}, which is the pioneering work to extract features directly on raw point clouds. It is widely deployed as the feature extractor \citep{wang2021unsupervised,poursaeed2020self,sauder2019self} due to its simple and lightweight network structure. Taking advantage of the point permutation invariance, PointNet aligns the input points to a canonical space and aggregates global features by symmetric functions such as max pooling.

However, it fails to capture local structures induced by the metric space in which the points reside, thereby limiting its ability to recognize fine-grained patterns and generalize to complex scenes. The updated version PointNet++ \citep{qi2017pointnet++} was then put forward several months later. It adopts multi-scale, multi-resolution sampling, and groping strategies to propagate features from one level to another, which improves the feature learning ability further. Furthermore, the point patch generation strategy combining Farthest Point Sampling (FPS) and K-Nearest Neighbor (KNN) provides a template for point cloud cropping preprocessing for subsequent studies \citep{yu2021point,pang2022masked,zhang2022masked}.

\subsubsection{VoxelNet}
VoxelNet \citep{YinZhou2018VoxelNetEL} is a generic point-specific network that uses voxels (i.e. finite unit cubes), to divide and access a local representation of point clouds for 3D detection tasks \citep{li2022deepfusion,min2022voxel,hess2022masked}. This network partitions disordered point clouds and performs feature learning in quantified and fixed-size 3D structures. One innovation is the stacking Voxel Feature Encoding (VFE) layers which encode interaction between points within a voxel and grasp descriptive appearance information. The output of each VFE layer is the concatenation of point-wise features and locally aggregated features so that local features are better captured. However, the expensive computation of voxel construction and quantization artifacts constrain the model from capturing high-resolution or fine-grained representations.

\subsubsection{DGCNN}
A point with its neighbors can reflect the geometry property of a local point cloud. Such a local relationship could be expressed by a graph network. Therefore, Wang et al. proposed a dynamic graph-based CNN network (DGCNN) \citep{wang2019dynamic} that encodes the edge features between vertices. Instead of learning point representations directly, DGCNN represents the interactions between points and their edges in both Euclidean and semantic space, and learns the graph structure dynamically. This graph network-based architecture has served as a backbone in many subsequent point cloud SSL models with notable results \citep{poursaeed2020self,sauder2019self,afham2022crosspoint}.

\subsubsection{GAN}
Generative Adversarial Network (GAN) \citep{goodfellow2014generative} is a widely used framework in reconstruction-based pretext tasks for point cloud knowledge mining. It consists of two components: the generator, which generates point clouds similar to the training data, and the discriminator, which distinguishes between generated and real points. These two modules are trained under an adversarial paradigm without supervision. The framework can be formulated as a two-player minimax game:

\begin{equation}
    \mathop{min}\limits_{G}\mathop{max}\limits_{D}E_{x\in X}[log(D(x))]+E_{z\in Z}[log(1-D(G(z)))]
\end{equation}

where $D$ and $G$ denotes the discriminator and the generator, and $X$ and $Z$ represent the data and noise distribution, respectively.

\subsubsection{Transformers}
Transformers have become one of the most prevalent architectures in many fields. They benefit from the multi-head self-attention mechanism, which allows them to capture long-range dependencies between point patches and discover implicit regional correlations. The state-of-the-art performance on SSL point cloud classification and part segmentation has been achieved by transformer-based models such as the one proposed by Zhang et al. \citep{zhang2022masked}. Furthermore, point cloud transformer (PCT) \citep{guo2021pct}, a variant adapted specifically for point clouds, enhances local feature extraction with the support of farther point sampling and nearest neighbor search, and further improves performance on various downstream tasks.

\subsection{Pseudo labels}
Pseudo labels are introduced in point cloud SSL due to the absence of ground truth labels. It facilitates the calculation of loss with the output of the pretext tasks, which is then used for updating encoders via backpropagation. Information contained in pseudo labels is often considered as a more reliable and informative source for pretext tasks to learn point cloud representation than tags. For instance, the label 'airplane' only indicates the shape of objects without descriptions like colors, poses, and differences from other samples in same category. In contrary, these attributes are implicitly contained in point clouds and could be captured as pseudo label in SSL tasks.

Different methods define pseudo labels in different ways. In most reconstruction-based pretext tasks, pseudo labels are point cloud itself which provides a rebuilding objective for pretext task. In contrast-based methods, pseudo labels are a multidimensional matrix carrying collection information and are typically generated using clustering methods such as memory bank \citep{wu2018unsupervised}, online dictionary \citep{he2020momentum}, and prototype approaches \citep{caron2020unsupervised}, representing mean and variance of all or part of the features of point cloud dataset. For some alignment-based prediction or motion-based tasks pre-trained on temporal point cloud datasets, pseudo labels are geometric information like position, pose, and orientation in a number of frames before and after.

\subsection{Loss functions}
Appropriate and easily-differentiable loss functions are critical to facilitate backpropagation and optimization for encoders. In reconstruction-based pretext tasks, the symmetric function, Chamfer distance (CD), is commonly employed to assess the distance between each point in one set and its corresponding nearest point in the other. More formally, for two non-empty subsets $X$ and $Y$, Chamfer distance $d_{CD}(X,Y)$ is defined as:

\begin{equation}
    d_{CD}(X,Y) = \frac{1}{|X|} \sum_{x \in X} \min_{y \in Y} ||x-y||^2 + \frac{1}{|Y|} \sum_{y \in Y} \min_{x \in X} ||x-y||^2
\end{equation}

Here, $x$ and $y$ represent the points in the reconstruction point set $X$ and the original input point set $Y$, respectively; $|| \cdot ||$ denotes the L2 distance between two points and $| \cdot |$ refers to the number of points. The smaller the CD value, the more similar the two point sets are, and the better the SSL algorithm performs.

For contrast-based pretext tasks, the objective is to discriminate the similarities and differences between each point cloud samples on the overall semantic level. A cross-entropy like loss function to encourage the positive samples to be close to each other (and negative ones to be far from each other) is needed. InfoNCE (NCE stands for Noise-Contrastive Estimation) is a contrastive loss function that estimates the mutual information between a pair of samples, and can be formulated as:

\begin{equation}
    L_q = - \log \frac{\exp(q \cdot k_+ / \tau)}{\sum_{i=0}^K \exp(q \cdot k_i / \tau)}
\end{equation}

where $q$ indicates the encoded query (feature); $k$ indicates a set of $K+1$ encoded samples $\{ k_0, k_1, k_2, \dots, k_K \}$, which could be regarded as the prototypes of historical samples; $\tau$ is the temperature parameter controlling the sharpness of the distribution. Assuming there is only one positive sample $k_+$ in the set $k$ matching the query $q$, the others $K$ samples are all negative. InfoNCE aims to assign the query $q$ into the positive sample $k_+$ in the $K+1$ classification problem \citep{he2020momentum}. In other words, the loss function tries to maximize the logits of $q \cdot k_+$ and minimize the value of the denominator.

\section{Self-Supervised Learning pretext tasks for point cloud}\label{sec:pretext}
We classify the current point cloud SSL research into four general categories based on the nature of the pretext tasks: reconstruction-based, contrast-based, alignment-based, and motion-based methods, as shown in Fig. \ref{fig:taxonomy}. These categories can be further divided into more fine-grained sub-categories according to the different ways in which the features are extracted and used. The following sections summarize the principles and peculiarities of various proxy tasks in details. It should be noted that some methods may reside in multiple sub-categories.

\subsection{Reconstruction-based methods}
Reconstruction-based methods learn point cloud representations by reconstructing the corrupted point clouds and recovering the original ones as much as possible. Global features as well as the mappings between local and global areas are learned during the reconstruction process. According to different types of corruption and reconstruction objects, we further divide them into six sub-categories: mask recovery, spatial restoration, point sampling, disentanglement, deformation reconstruction, and generation and discrimination. Summary about the methods under these six sub-categories is shown in Table \ref{tab:reconstruction}.

\begin{table*}[]
    \centering
    \caption{Summary of reconstruction-based point cloud SSL methods. D \& G stand for Generation and Discrimination.}
    \label{tab:reconstruction}
    \resizebox*{\linewidth}{!}{
    \begin{tabular}{|c|c|c|c|}\hline
        Year & Method & Sub-categories & Contributions  \\\hline
        2021 & Point-BERT \citep{yu2021point} & Mask recovery & Reconstructing missing point tokens with BERT-style transformer \\
        2022 & Point-MAE \citep{pang2022masked} & Mask recovery & Shifting masked tokens to decoder to avoid early leakage \\
        2022 & MaskSurf \citep{zhang2022masked} & Mask recovery & Estimating surfel position and per-surfel orientation simultaneously \\
        2022 & Voxel-MAE \citep{min2022voxel} & Mask recovery & Performing additional binary voxel classification for complicated semantics awareness \\\hline
        2019 & 3D jigsaw \citep{sauder2019self} & Spatial restoration & Rearranging randomly disorganized point clouds \\
        2019 & CloudContext \citep{sauder2019context} & Spatial restoration & Predicting relative structural position between two given patches \\
        2020 & Orientation estimation \citep{poursaeed2020self} & Spatial restoration & Predicting and recovering rotation angle around an axis \\\hline
        2019 & PU-GAN \citep{li2019pu} & Point upsampling / D \& G & Utilizing self-attention unit for feature aggregation and quality enhancement \\
        2021 & SSPU-Net \citep{zhao2021sspu} & Point upsampling & Leveraging shape coherence between sparse input and generated dense point cloud \\
        2022 & UAE \citep{zhang2022upsampling} & Point upsampling & Gaining both advanced semantic information and basic geometric structure \\
        2022 & SPU-Net \citep{liu2022spu} & Point upsampling & Integrating self-attention with graph convolution network for context feature extraction \\
        2022 & PUFA-GAN \citep{liu2022pufa} & Point upsampling / D \& G & Employing graph filter to extract high frequency points of sharp edges and corners \\
        2022 & SSAS \citep{zhao2022self} & Point upsampling & Achieving magnification-flexible point clouds upsampling \\\hline
        2022 & Pose Disentanglement \citep{tsai2022self} & Disentanglement & Uncoupling content and pose attributes in partial point clouds \\
        2022 & CP-Net \citep{xu2022cp} & Disentanglement & Disentangling point clouds into contour and content ingredients \\
        2022 & MD \citep{sun2022self} & Disentanglement & Separating mixing point cloud into two independent objects \\\hline
        2018 & FoldingNet \citep{yang2018foldingnet} & Deformation reconstruction & Stretching 2D grid lattice to reproduce 3D surface structure \\
        2021 & Self-correction \citep{chen2021shape} & Deformation reconstruction & Recovering shape-disorganized point regions \\
        2021 & DefRec \citep{achituve2021self} & Deformation reconstruction & Performing deformation on 2D grids to fit arbitrary 3D object surface \\\hline
        2018 & PC-GAN \citep{li2018point} & D \& G & Employing hierarchical and interpretable sampling strategy \\
        2019 & RL-GAN \citep{sarmad2019rl} & D \& G & Introducing reinforcement learning agent to control GAN \\
        2019 & TreeGCN \citep{shu20193d} & D \& G & Leveraging ancestor information to boost point representation \\
        2022 & MaskPoint \citep{liu2022masked} & D \& G & Performing simple binary classification as proxy task \\\hline
    \end{tabular}}
\end{table*}

\subsubsection{Mask recovery}

\begin{figure}[htbp]
    \centering
    \includegraphics[width=0.97\linewidth]{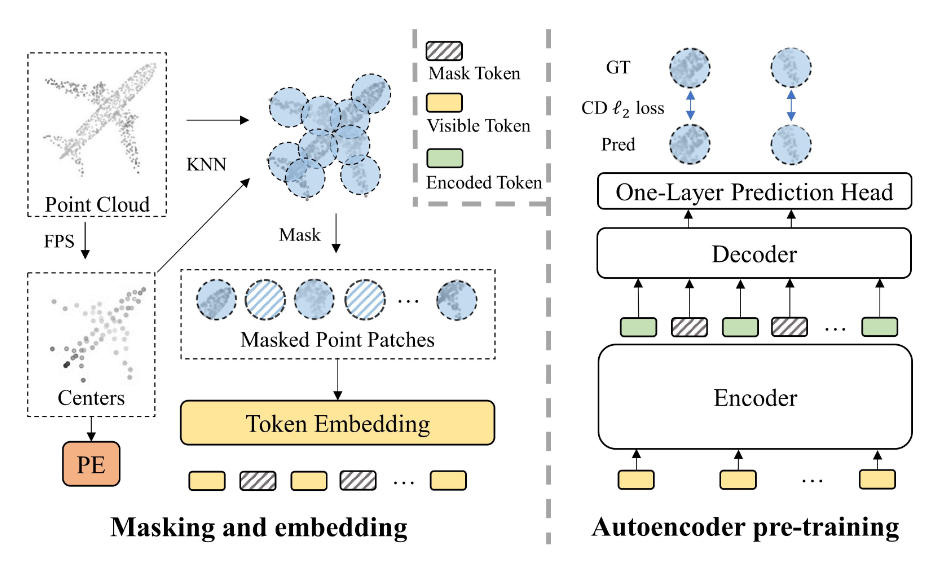}
    \caption{The general pipeline of Point-MAE. (1) The process of masking and embedding is demonstrated on the left. The point cloud patches are generated by FPS and KNN and masked randomly. Both visible and mask patches are mapped to the corresponding tokens through PointNet-based embedding layers. Also, the Position Embedding (PE) is obtained by mapping the center coordinates to the embedding dimension. (2) The autoencoder pre-training is shown on the right. The encoder only processes the visible tokens while the mask tokens are shifted and added to the input sequence of the decoder to reconstruct the masked patches. This figure is adapted from \citep{pang2022masked}.}
    \label{fig:Point-MAE}
\end{figure}

The core idea of reconstruction is to mask a portion(s) of the point cloud and recover such missing part via an encoder-decoder architecture. Similar to the image inpainting task \citep{sarmad2019rl} and Mask AutoEncoder (MAE) \citep{hess2022masked}, the encoder is required to capture the local geometric structure and the regional relations during the restoration process. Generally speaking, the better the reconstruction, the more effective the learned features.

Point-BERT \citep{yu2021point}, built based on BERT \citep{devlin2018bert}, designs a point-specific tokenizer on discrete Variational AutoEncoder (dVAE) to map patches to discrete tokens to capture meaningful local geometric patterns. A portion of the input is randomly masked out, and a BERT-style transformer is trained to reconstruct the missing token under the supervision of point tokens obtained by the tokenizer. However, the tokenizer should be pre-trained in advance, and Point-BERT over-relies on auxiliary contrastive learning as well as data augmentation. 

To address this issue, Pang et al. proposed Point-MAE \citep{pang2022masked} as a neat and efficient scheme of mask autoencoder as shown in Fig. \ref{fig:Point-MAE}. Concretely, Point-MAE employs the standard transformer as the backbone with an asymmetric encoder-decoder architecture to process random masking points with a high ratio (60\%-80\%). The mask tokens are shifted from the input of the encoder to the lightweight decoder, which saves considerable computation, and more significantly, avoids early leakage of location information. To further capture local geometric information, Zhang et al. introduced Mask Surfel Prediction (MaskSurf) \citep{zhang2022masked}, which estimates the surfel position (i.e., points) and per-surfel orientation (i.e., normals) simultaneously. Such a two-head pre-training paradigm has been justified to capture more effective representations than a reconstruction-only pretext. Likewise, Voxel-MAE \citep{min2022voxel} transforms point clouds into volumetric representations and applies the range-aware random masking strategy on the voxel grid. Besides reconstructing the occupancy value of masked voxels, a supplementary binary voxel classification task distinguishing whether the voxel contains point clouds boosts the model to learn more complicated semantics.

\subsubsection{Spatial restoration}

\begin{figure}[htbp]
    \centering
    \includegraphics[width=0.97\linewidth]{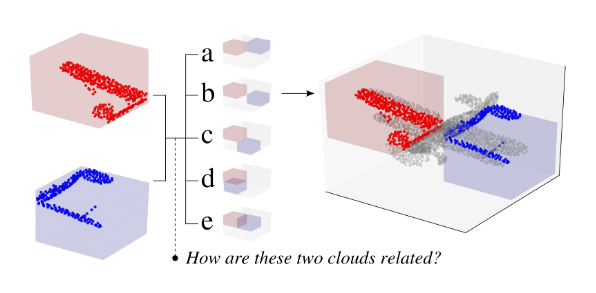}
    \caption{Illustration of the CloudContext pretext task. The pre-training model is enforced to estimate the spatial relevance between two given point cloud segments from six categories. In this case, the exact relation of these two components is 'the red part is diagonally above the blue part'. This figure is adapted from \citep{sauder2019context}.}
    \label{fig:CloudContext}
\end{figure}

Point clouds are the coordinate sets containing abundant spatial information that describes the structural distribution of objects and the environment in the Euclidean space. It is natural to exploit such rich spatial knowledge as the supervision signal in pretext tasks.

Sauder et al. \citep{sauder2019self} proposed a 3D version of the jigsaw pretext to rearrange point clouds whose parts have been randomly disrupted and displayed by voxels along the axes. The goal of this pretext is to restore the original position of each patch (labeled by voxel ID) from the state of chaotic and disorderly distribution. They later developed CloudContext \citep{sauder2019context} to forecast the spatial relevance between two point cloud segments. As shown in Fig. \ref{fig:CloudContext}, the model is trained to predict the relative structural position between two given patches from the same object, which utilizes the innate attributes of point clouds as they are not restrained by a discrete grid. By doing so, powerful per-point features can be accessed in an easy-to-implement unsupervised manner without expensive computation. 

Orientation estimation \citep{poursaeed2020self} is another simple but effective proxy task to capture the spatial information of point clouds. With the canonical orientation provided in most datasets, the orientation estimation pretext task aims to predict and recover the rotation angle around an axis via matrix multiplication. Such a pretext requires a high-level holistic understanding of shapes and obviates the need for manual annotations.

\subsubsection{Point upsampling}
\begin{figure}[htbp]
    \centering
    \includegraphics[width=0.97\linewidth]{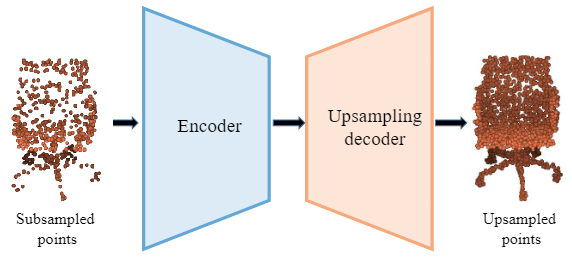}
    \caption{Overview of Upsampling AutoEncoder. The input point cloud is subsampled by a random sampling strategy and then fed into the encoder to extract point-wise features. The decoder is adopted to reconstruct the original point cloud with offset attention based on the learned representation. This figure is adapted from \citep{zhang2022upsampling}.}
    \label{fig:UAE}
\end{figure}

Point upsampling is the operation to upsample sparse, noisy, and non-uniform point clouds to generate a dense, complete, and high-resolution point cloud, which is challenging but also beneficial for the model to capture implicit geometric representations of the underlying surface.

PU-GAN \citep{li2019pu} is a pioneer SSL upsampling paradigm formulated based on the generative adversarial network (GAN) \citep{goodfellow2014generative} to generate a diverse range of point distributions from the latent space and upsample points over patches. An up-down-up unit is embedded in the generator to expand point features as well as a self-attention unit for quality enhancement on feature aggregation. The discriminator is inspired to gain inherent patterns and improve the uniformity of output generation according to a compound loss including adversarial, uniform, and reconstruction terms. Motivated by PU-GAN, Zhang et al. proposed the Upsampling AutoEncoder (UAE) \citep{zhang2022upsampling} to gain both advanced semantic information and basic geometric structure from subsampled point clouds. As shown in Fig. \ref{fig:UAE}, the encoder is devised to perform point-wise feature extraction on the subsampled point cloud, and the upsampling decoder is designed to reconstruct the original dense point cloud with offset attention \citep{guo2021pct} to refine global shape structure. 

Liu et al. \citep{liu2022spu} proposed a coarse-to-fine reconstruction framework, dubbed SPU-Net, integrating self-attention with graph convolution network (GCN) for contextual feature extraction and generating fine point sets with hierarchically learnable 2D grids. Zhao et al. \citep{zhao2021sspu} introduced SSPU-Net by leveraging the shape coherence between input sparse and generated dense point clouds. In addition, it has an image-consistent loss among multi-view rendered images to capture the latent patterns of underlying point structures. 

PUFA-GAN \citep{liu2022pufa}, a frequency-aware framework, utilizes a graph filter to extract high frequency (HF) points of sharp edges and corners so that the discriminator could focus on the HF geometric properties and enforce the generator producing neat and more uniform upsampled point clouds. To get rid of the fixed upsampling factor restriction, Zhao et al. \citep{zhao2022self} presented a self-supervised arbitrary-scale (SSAS) framework with a magnification-flexible upsampling strategy. Instead of direct mapping from sparse to dense point clouds, the proposed scheme seeks the nearest projection points on the implicit surface for seed points via two functions, which are exploited to estimate the projection direction and distance, respectively.

\subsubsection{Disentanglement}
\begin{figure}[htbp]
    \centering
    \includegraphics[width=0.97\linewidth]{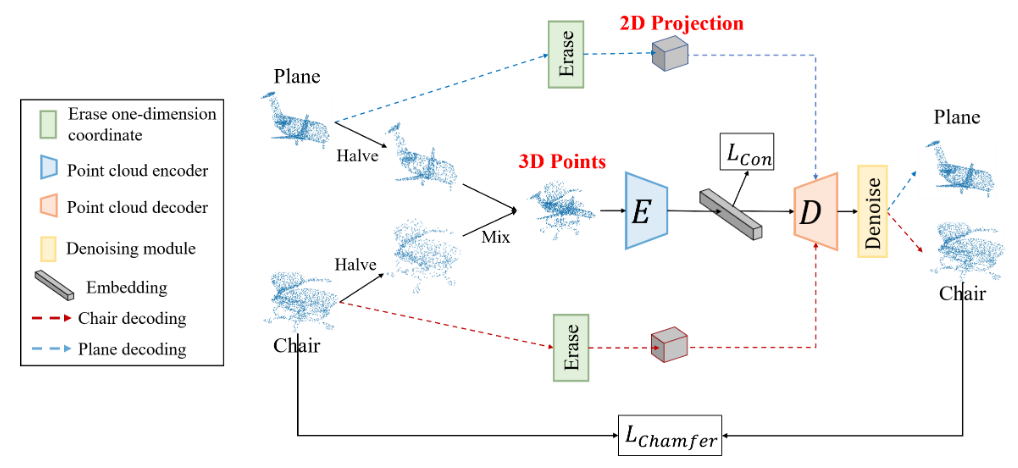}
    \caption{The schematic of Mixing and Disentangling (MD) pretext. The two input point clouds are separately halved and mixed into a hybrid object feeding to the encoder $E$ to mine the geometry-aware embedding. The 'Erase' operation is applied to obtain the 2D projection from both original input point clouds simultaneously. The instance-adaptive decoder $D$ receives the embedding together with the two partial projections as input to disentangle the blended shape into the original two point clouds. The chamfer distance is used to measure the reconstruction error between generated point clouds and the original ones. This figure is adapted from \citep{sun2022self}.}   
    \label{fig:disentanglement}
\end{figure}

Models pre-trained under the SSL paradigm usually tend to learn well the low-level geometric features of point clouds, such as pose, contour, and shape information, but overlook the high-level semantic content understanding, which often leads to unsatisfactory performance in downstream tasks such as object classification that requires global discriminative capability. To tackle this issue, disentanglement-based SSL pretexts are proposed to separate the low-level geometric features from the high-level semantic embedding. Feature extraction is performed based on various contents using distinct modules to obtain hierarchical representations.

Tsai et al. \citep{tsai2022self} proposed a disentanglement framework that uncouples content and pose attributes in partial point clouds to enhance both geometric and semantic feature abstraction. Two encoders are employed to learn the content and multi-view poses separately, where the gained pose representation should predict the viewing angle and navigate the partial point cloud reconstruction cooperated with the content from another specific view. Likewise, Xu et al. \citep{xu2022cp} presented a universal Contour-Perturbed Reconstruction Network (CP-Net) that disentangles a point cloud into contour and content ingredients. A concise contour-perturbed augmentation unit is exploited on the contour component and retains the content part of the point cloud. Therefore, the self-supervisor is able to concatenate the content component for advanced semantic comprehension.

Different from the above two pretexts, Mixing and Disentangling (MD) \citep{sun2022self} blends two disparate point shapes into a hybrid object and attains geometry-aware embedding from the encoder. An instance-adaptive decoder is then leveraged to restore the original geometries based on the obtained embedding by disentangling the mixed shape. As shown in Fig. \ref{fig:disentanglement}, except for the main encoder-decoder structure, the proposed scheme also encompasses a coordinate extracting operation 'Erase', which randomly drops one-dimension coordinate of each point to provide an extra 2D partial projection to better reconstruct the original point cloud shapes.

\subsubsection{Deformation reconstruction}

\begin{figure}[htbp]
    \centering
    \includegraphics[width=0.97\linewidth]{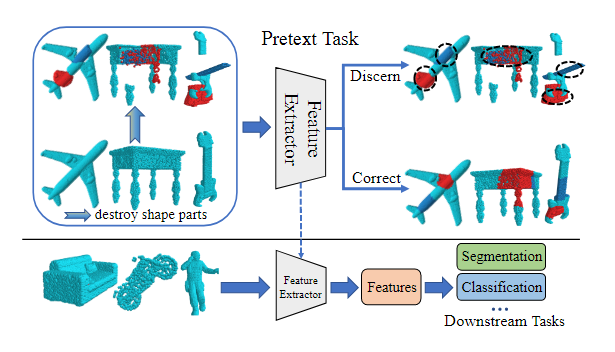}
    \caption{Demonstration of shape self-correction pretext. The input point cloud is firstly preprocessed by a shape-disorganizing module to generate a deformed point cloud and then fed to the encoder to learn the geometry-aware representation. Two separate task heads are constructed to distinguish and segment points belonging to distorted parts, and subsequently reconstruct the partial-deformed objects. The well-trained feature extractor is transferred to downstream tasks to estimate the feature capturing capability. This figure is adapted from \citep{chen2021shape}.}   
    \label{fig:self_correction}
\end{figure}

Point cloud deformation is a common phenomenon in real-world data scanning, which is usually caused by object distortion, sensor noise, or external occlusion. It has been discovered that SSL by reconstructing the original point cloud from the artificially deformed one (e.g. adding Gaussian noise or local translation) enables the learned model to obtain geometric perception as well as context awareness.

Chen et al. \citep{chen2021shape} proposed a shape self-correction pretext to mine implicit geometric embeddings of point clouds. The pretext assumes that a robust shape representation could identify and correct distorted regions of a shape. As shown in Fig. \ref{fig:self_correction}, the proposed scheme imposes destruction over certain regions by a shape-disorganizing module and sends the deformed point cloud to the feature extractor for embedding learning. Two task heads are built separately to discern the distorted components and further restore them to their original normal shapes for fine-grained geometric and contextual feature exploration.

Achituve et al. \citep{achituve2021self} conducted the first study of SSL for domain adaptation (DA) on point cloud via Deformation Reconstruction (DefRec). By mapping the dislocating points to their original location, the model is able to obtain the latent statistical structure of the input point cloud. Moreover, the distribution gap between source and target domains is bridged by the learned representation since they are invariant to distribution shift.

FoldingNet \citep{yang2018foldingnet} presents a novel folding-based decoder to perform deformation on the canonical 2D grid to fit an arbitrary 3D object surface. Instead of deforming the point cloud, the folding operation exerts a virtual force induced by the embedding captured from input to stretch a 2D grid lattice to reproduce the 3D surface structure. This approach tackles issues caused by point cloud's irregular attributes by applying implicit 2D grid constraints.

\subsubsection{Generation and discrimination}
\begin{figure}[htbp]
    \centering
    \includegraphics[width=0.97\linewidth]{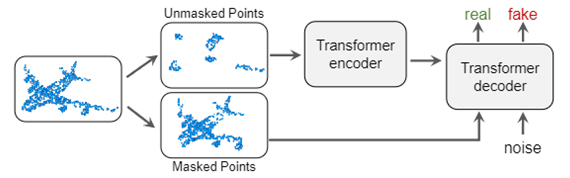}
    \caption{The general pipeline of the Mask-Point model. The reconstruction challenge is formulated as a discriminative pretext to determine whether the source of the extracted sample is a masked point cloud or a random noise. The figure is adapted from \citep{liu2022masked}.}
    \label{fig:maskpoint}
\end{figure}

The generation and discrimination pretext is a unique paradigm that designs a discriminator module to distinguish whether the fed point cloud is reconstructed from noise distribution or truly sampled. During the adversarial training process, the generator (encoder) and discriminator (decoder) compete with each other and are updated alternatively so that both components can be transferred for downstream tasks.

PC-GAN \citep{li2018point} is specifically designed for point clouds and employs a hierarchical and interpretable sampling strategy inspired by Bayesian and implicit generative models to tackle the issue of missing constraints on the discriminator. Sarmad et al. \citep{sarmad2019rl} introduced a reinforcement learning (RL) agent to control the GAN to extract implicit representations from noisy and partial input to generate high-fidelity and entire point clouds. Meanwhile, applying an RL agent to seek the best-fit input of GAN to produce low-dimensional latent embedding relieves the challenge of unstable GAN training. Shu et al. \citep{shu20193d} introduced a tree-structured graph convolutional network (TreeGCN) as the generator, leveraging ancestor information to boost the representation of the point. It is more efficient in computation than using neighborhood features as adopted in regular GCNs. PU-GAN \citep{li2019pu} and PUFA-GAN \citep{liu2022pufa}, both employed GANs-based models to generate dense and uniform point clouds with innovative modules for feature aggregation enhancement and high-frequency point filtering.

Liu et al. \citep{liu2022masked} proposed a discriminative mask pretraining transformer framework, MaskPoint, which combines mask and discrimination techniques to perform simple binary classification between masked object points and sampled noise. As shown in Fig. \ref{fig:maskpoint}, the original complete point cloud is divided into 90\% masking portion and a 10\% visible potion. Two kinds of query, where the real is sampled from masked point clouds while the fake is derived from random noise, are fed to the decoder for classification. During the discrimination process, the model is required to deduce the full geometry from small visible portions.

\subsection{Contrast-based methods}
Contrastive learning is a popular mode of SSL that encourages augmentation of the same input to have more comparable representations. The general approach is to expand the views of input point clouds (anchors) by various data augmentation techniques. In particular, it tries to enforce positive samples augmented from the same anchor more similar than negative samples from different anchors in the feature space. In this section, we will introduce contrast-based methods with representative examples and discuss their contributions and limitations. A brief summary of these methods is shown in Table \ref{tab:contrast}.

\begin{table*}[]
    \centering
    \caption{Summary of contrast-based point cloud SSL methods.}
    \label{tab:contrast}
    \resizebox*{\linewidth}{!}{
    \begin{tabular}{|c|c|c|c|}\hline
        Year & Method & Sub-categories & Contributions  \\\hline
        2020 & Info3D \citep{sanghi2020info3d} & Object contrast & Maximizing mutual information between objects and their transformations \\
        2022 & AFSRL \citep{lu2022joint} & Object contrast & Imposing data-level augmentation and feature enhancement simultaneously \\
        2019 & Contrasting and clustering \citep{zhang2019unsupervised} & Object contrast & Solving part contrast and object cluster tasks consecutively \\
        2021 & Hard negatives \citep{du2021self} & Object contrast & Leveraging self-similar point cloud patches; facilitating hierarchical context primitives capturing \\\hline
        2020 & PointContrast \citep{xie2020pointcontrast} & Scene contrast & Obtaining dense features at point-level on complex scenes by point contrast \\
        2021 & Contrastive Scene Contexts \citep{hou2021exploring} & Scene contrast & Introducing ShapeContext local descriptor and achieving data-efficiency \\
        2021 & CoCoNets \citep{lal2021coconets} & Scene contrast & Mapping RGB-D images to 3D points by optimizing view-contrastive prediction \\
        2020 & P4Contrast \citep{2020arXiv201213089L} & Scene contrast & Utilizing synergies between two modalities for better feature extraction \\
        2021 & DepthConstrast \citep{zhang2021self} & Scene contrast & Applying Instance Discrimination on depth maps \\\hline
    \end{tabular}}
\end{table*}

\subsubsection{Object contrast}

\begin{figure}[htbp]
    \centering
    \includegraphics[width=0.97\linewidth]{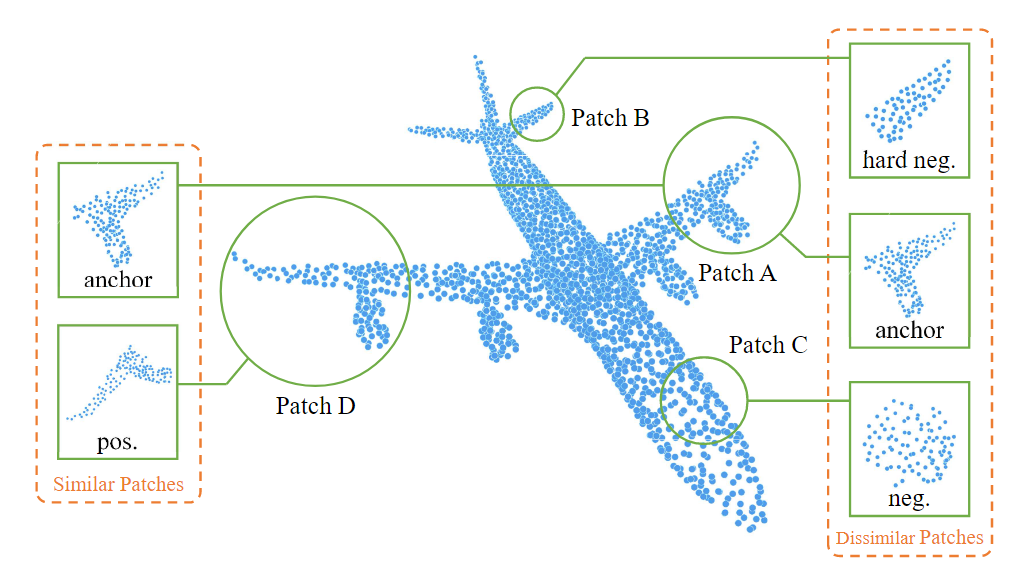}
    \caption{Illustration of a self-contrastive paradigm. Patch A is selected as the anchor and the symmetrical part Patch D is the positive sample. Patch B and C are the negative samples, where Patch B is hard to distinguish due to its comparative similarity to the anchor. The figure is adapted from \citep{du2021self}.}
    \label{fig:hard_negative_sampling}
\end{figure}

Traditional contrastive learning research usually focuses on instance-wise objects. The priority is on overall semantic learning through discriminative pretext tasks that capture context similarity and difference of point clouds. Such an object-contrast paradigm performs data augmentation on relatively large patches or whole single point objects to capture global geometric awareness.

Sanghi \citep{sanghi2020info3d} proposed Info3D, which takes inspiration from Contrastive Predictive Coding \citep{oord2018representation} and Deep InfoMax \citep{velickovic2019deep}, to obtain rotation-insensitive representation by maximizing mutual information between 3D objects and their local chunks as well as geometrically transformed versions. Lu et al. \citep{lu2022joint} proposed the Augmentation Fusion Self-Supervised Representation Learning (AFSRL) framework, which imposes data-level augmentation and feature enhancement simultaneously to construct a stable and invariant point cloud embedding. The correspondence between augmented pairs is acquired, and the invariant semantic is maintained under perturbations during augmentation. 

Zhang et al. \citep{zhang2019unsupervised} introduced a simple two-phase unsupervised GCN framework (contrasting and clustering), to capture superior point embedding by solving part contrast and object cluster tasks consecutively. Du \citep{du2021self} presented a self-contrastive paradigm leveraging self-similar point cloud patches within a single point cloud to facilitate local shape and global context primitives capturing. As shown in Fig. \ref{fig:hard_negative_sampling}, according to the nonlocal self-similar property of the point cloud, where regional geometry remains invariant after affine transformation, the self-similar point cloud patches are treated as positive samples otherwise negative based on the inferred similarity score. Moreover, hard negative samples, close to positive samples in the representation space, are sampled for more discriminative and expressive representation learning.

\subsubsection{Scene contrast}

\begin{figure}[htbp]
    \centering
    \includegraphics[width=0.97\linewidth]{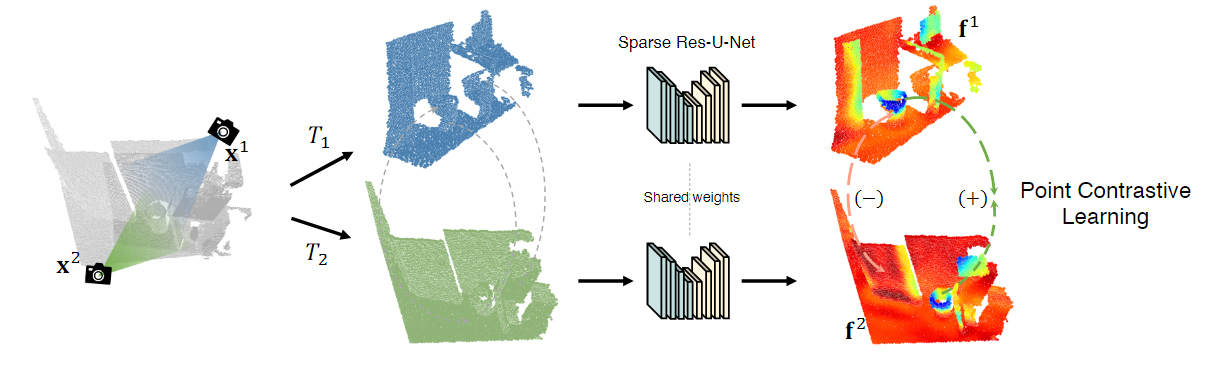}
    \caption{The illustration of PointContrast. Contrast is performed at the point-level between two transformed point clouds, where positive samples are the matched points while negative samples are the unmatched points across two views. The figure is adapted from \citep{xie2020pointcontrast}.}
    \label{fig:pointcontrast}
\end{figure}

Different from object-contrast, the scene-contrast paradigm concentrates on scenes to capture broader environmental context and neighborhood perception, which is more relevant to real-world complex scenarios.

To address the domain gap issue (i.e., it is insufficient to capture a global representation from object instances), Xie et al. \citep{xie2020pointcontrast} proposed PointContrast, a sparse residual U-Net based framework aiming to obtain dense features at the point-level on complex scenes. As shown in Fig. \ref{fig:pointcontrast}, two views $x^1$ and $x^2$ are produced from a complicated point cloud scene, where corresponding pairs are computed between these two views as the positive samples. Two rigid transformations $T^1$ and $T^2$ are utilized to increase the difficulty of the pretext which demands the network to learn the invariant embedding under random geometric shift. The contrastive loss is defined to shorten the distance between the matched points and enlarge the distance of mismatched points of the two overlapping partial scans so that the pre-training model can capture local descriptions and be universally pertinent to various advanced 3D understanding downstream tasks.

However, PointContrast only considers point correspondence matching but ignored the spatial configurations and contexts in a scene, e.g., relative pose and distance, therefore confining its transferability and scalability. To address this issue, Hou et al. \citep{hou2021exploring} presented Contrastive Scene Contexts to fuse spatial information into pre-training objects by introducing ShapeContext local descriptor \citep{xie2018attentional} partitioning and performing contrastive learning in each region. The method improves the performance and data efficiency on downstream tasks in which employing only 0.1\% of point labels reaches the performance level with full supervision.

Continuous Contrastive 3D Networks (CoCoNets) \citep{lal2021coconets} aims to infer latent scene representations by mapping RGB-D images to 3D point scenarios and optimizing view-contrastive prediction. P4Contrast \citep{2020arXiv201213089L}, another RGB-D bi-modal SSL framework, proposes to contrast point-pixel pairs and provides additional flexibility for hard negative creation to exploit the synergies between two modalities for better feature extraction. DepthConstrast \citep{zhang2021self} circumvents the need for point correspondences and instead applies the Instance Discrimination \citep{wu2018unsupervised} method on depth maps combined with a momentum encoder to improve the geometric perception.

\subsection{Alignment-based methods}
Point cloud representation is generally invariant to transformations in terms of time flow, spatial motion, multi-view photography, etc. Based on this property, alignment-based methods have been proposed to learn the implicit embedding of point clouds by preserving the coherence of point features in spatiotemporal consistency, multi-view alignment, and multimodal fusion. A brief summary of the methods under this category is provided in Table \ref{tab:alignment}.

\begin{table*}[]
    \centering
    \caption{Summary of alignment-based point cloud SSL methods.}
    \label{tab:alignment}
    \resizebox*{\linewidth}{!}{
    \begin{tabular}{|c|c|c|c|}\hline
        Year & Method & Sub-categories & Contributions  \\\hline
        2020 & Info3D \citep{sanghi2020info3d} & Multi-view alignment & Maximizing mutual information between objects and their transformations \\
        2021 & OcCo \citep{wang2021unsupervised} & Multi-view alignment & Shielding and restoring occluded points in camera view \\
        2021 & Multi-view stereo \citep{yang2021self} & Multi-view alignment & Generating prime depth map as self-supervision signal \\
        2021 & Cross-view \citep{jing2021self} & Multi-view alignment & Jointly learning both 3D point cloud and 2D image embedding concurrently \\
        2022 & Multi-view rendering \citep{tran2022self} & Multi-view alignment & Encouraging 2D-3D global feature distributions to be similar \\\hline
        2021 & Order prediction \citep{wang2021self} & Spatiotemporal consistency & Sorting temporal order of sampled and disorganized point cloud clips \\
        2021 & STRL \citep{huang2021spatio} & Spatiotemporal consistency & Dual-branch network to predict representation of another temporally correlated input \\
        2022 & Futrue prediction \citep{mersch2022self} & Spatiotemporal consistency & Forecasting future point cloud scenes with lightweight model \\\hline
        2020 & PointPainting \citep{vora2020pointpainting} & Multimodal fusion & Projecting LiDAR points into semantic segmentation diagram for traffic scenes \\
        2021 & PointAugmenting \citep{wang2021pointaugmenting} & Multimodal fusion & Replacing sub-optimal segmentation scores with high-dimension CNN features \\
        2022 & DeepFusion \citep{li2022deepfusion} & Multimodal fusion & Exploiting cross-attention to capture long-range correlations of image-LiDAR pairs \\\hline

    \end{tabular}}

\end{table*}

\subsubsection{Multi-view alignment}
\begin{figure}[htbp]
    \centering
    \includegraphics[width=0.97\linewidth]{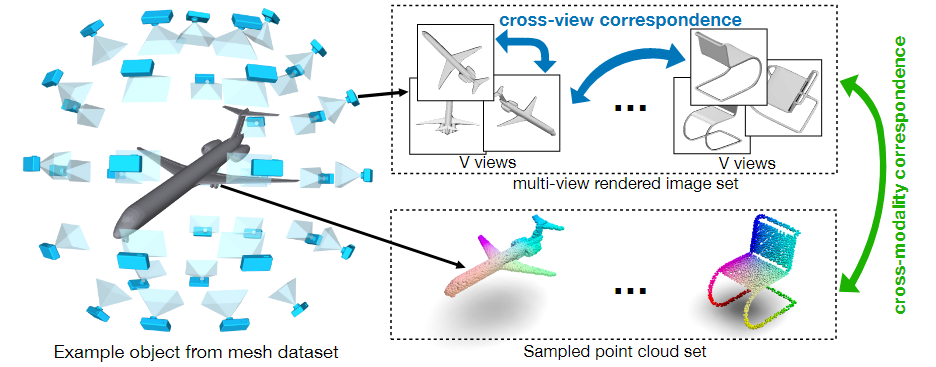}
    \caption{The schematic view of cross-modality and cross-view correspondences. The 3D point cloud objects and corresponding pairs of multi-view rendered images are sampled from the same mesh input, respectively. The relation of diverse views is captured as the supervision signal by sustaining the alignment among multi-view and cross-domain representations. The figure is adapted from \citep{jing2021self}.}
    \label{fig:cross_view}
\end{figure}

Compared to direct processing and feature extraction on 3D point clouds, projecting point clouds into 2D images for dimension reduction and utilizing mature image networks as well as 2D SSL technologies is relatively more accessible. To ensure that the learned embeddings sufficiently represent the entire 3D point cloud objects or scenes, multi-view alignment pretexts are necessary to preserve the integrity and uniformity of the point cloud features.

Info3D \citep{sanghi2020info3d} aims to obtain rotation-insensitive representations by maximizing mutual information between 3D objects and their local chunks for patch-level consistency. Occlusion Completion (OcCo) \citep{wang2021unsupervised} combines the idea of mask recovery shielding and restoring occluded points in a camera view for better spatial and semantic properties comprehension. Similarly, Yang et al. \citep{yang2021self} introduced an SSL multi-view stereo structure generating prime depth map as pseudo-labels and refined such self-supervision from neighboring views as well as high-resolution images by multi-view depth fusion iteratively. Furthermore, the correspondence of pixel/point of the point clouds and the corresponding multi-view images are aligned for cross-modality consistency. 

Jing et al. \citep{jing2021self} proposed a novel SSL framework leveraging cross-modality and cross-view correspondences to jointly learn both 3D point cloud and 2D image embedding concurrently. As shown in Fig. \ref{fig:cross_view}, point cloud objects and comparable pairs of multi-view rendered images are sampled from the same mesh input. In addition to 2D-3D consistency, the contrastive notion is adopted into cross-view alignment that shortens intra-object distance while maximizing inter-object discrepancy of distinct rendered images. Similarly, Tran et al. \citep{tran2022self} presented a dual-branch model not only agreeing upon fine-grained pixel-point local representation but also encouraging 2D-3D global feature distributions as approaching as possible by exploiting knowledge distillation.

\subsubsection{Spatiotemporal consistency}
\begin{figure}[htbp]
    \centering
    \includegraphics[width=0.97\linewidth]{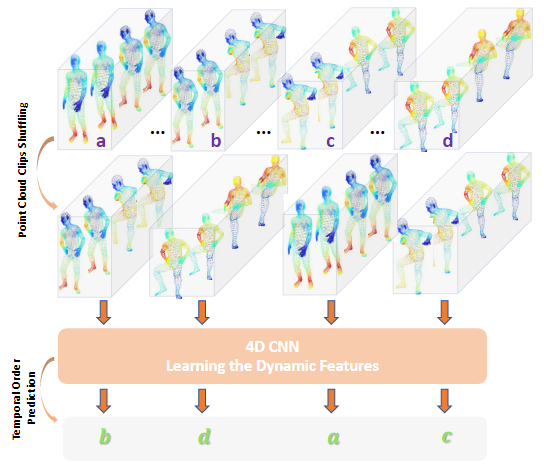}
    \caption{Demonstration of point cloud sequence order prediction. The first row is the uniformly sampled point cloud clips from the continuous point cloud sequence. Then these clips are randomly shuffled and then fed into 4D CNN in the second row to learn the dynamic features of human actions, The original temporal order is predicted in a self-supervised manner. The figure is adapted from \citep{wang2021self}.}  
    \label{fig:order_prediction}
\end{figure}

Unlike previous methods, the spatiotemporal approach is more concerned with long-range spatial and temporal invariance before and after certain point cloud frames, which are 4D data (XYZ coordinate + temporal dimension), to capture intrinsic characteristics of dynamic sequences.

Motivated by the success of Xu et al.'s work \citep{xu2019self} in video SSL, Wang et al. proposed the first SSL scheme to gain effective temporal embeddings on dynamic point cloud data by sorting the temporal order of sampled and disorganized point cloud clips. As shown in Fig. \ref{fig:order_prediction}, a few static point cloud frames are uniformly sampled and disordered, which are then processed by a 4D CNN to restore the disrupted fragments to the correct order on an unannotated, large-scale, sequential point cloud action recognition dataset. 

Another spatiotemporal representation learning (STRL) \citep{huang2021spatio} framework, inspired by BYOL \citep{grill2020bootstrap}, designs a dual-branch pipeline, referred to as online and target networks, to collaborate and promote each other. Specifically, the online network is enforced to predict the target network representation of another temporally correlated input, which is augmented by random spatial transformation, for spatiotemporal invariant contextual cues extraction. Taking training and inference time into account, Mersch et al. \citep{mersch2022self} presented an innovative 3D spatiotemporal convolution encoder-decoder neural network consisting of fewer parameters to predict future point cloud scenes. Such a lightweight model concatenates range images as input to estimate forthcoming images and per-point scores in multiple future steps, so that spatial and temporal scene information can be captured simultaneously.

\subsubsection{Multimodal fusion}

\begin{figure}[htbp]
    \centering
    \includegraphics[width=0.97\linewidth]{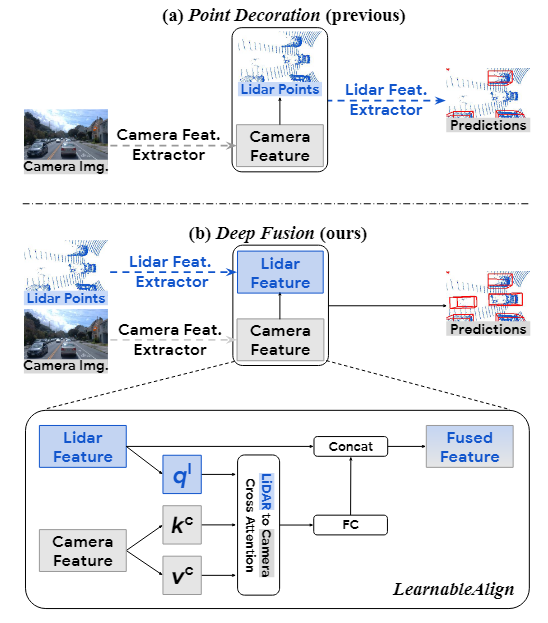}
    \caption{Demonstration of point decoration and deep fusion. (a) Previous cross-modal paradigms \citep{lal2021coconets,jing2021self} decorate LiDAR points with camera feature on input-level for 3D detection. (b) DeepFusion \citep{li2022deepfusion} fuses camera and LiDAR features extracted by respective encoders and leverages cross-attention consistency technique. The figure is adapted from \citep{li2022deepfusion}.}  
    \label{fig:deepfusion}
\end{figure}

Rather than simply requiring coherence between 2D-3D correspondences \citep{lal2021coconets,jing2021self,tran2022self}, automatic driving algorithms demand sophisticated collaboration between in-vehicle sensors. For example, cameras and LiDARs provide complementary information (e.g., colorful texture visualization and distance perception) for 3D object detection. Therefore, multimodal fusion is a promising direction to exploit the potential of images and point clouds for acquiring effective traffic scene features.

Vora et al. \citep{vora2020pointpainting}, Wang et al. \citep{wang2021pointaugmenting}, and Li et al. \citep{li2022deepfusion} offered compact frameworks for tight sensor-fusion which could be implemented under the SSL paradigm without human annotations. PointPainting \citep{vora2020pointpainting} is a sequential fusion method that projects LiDAR points onto semantic segmentation diagrams for traffic scenes with color marking. Each point is painted with a class score obtained from the image segmentation network and then can be utilized in any LiDAR detection approaches. Such a painting fusion method cleverly addresses the limitations of depth-blurring and scale ambiguity by consolidating the birds-eye and camera view. 

PointAugmenting \citep{wang2021pointaugmenting} adopts a late cross-modal fusion mechanism based on PointPainting, replacing the sub-optimal segmentation scores with high-dimension CNN features containing rich outlook hints and larger receptive fields to emphasize the delicate details. Moreover, a simple yet effective cross-modal data augmentation pastes virtual objects into images and point clouds for alignment between the camera and LiDAR. However, both PointPainting and PointAugmenting simply decorate LiDAR points with camera embeddings as shown in Fig. \ref{fig:deepfusion}(a). To improve the performance on downstream tasks, DeepFusion \citep{li2022deepfusion} proposed an end-to-end cross-modal fusion on the feature level, focusing on consistency improvement. As shown in Fig. \ref{fig:deepfusion}(b), a block named LearnableAlign is introduced to exploit cross-attention to dynamically capture long-range correlations during the image-LiDAR fusion process to enhance the model's recognition and localization capability.

\subsection{Motion-based methods}
Various point cloud frames contain rich geometric patterns and kinematic schemas that are concealed in the movement of objects or scenes. The motion-based SSL paradigm focuses on dynamically capturing the intrinsic motion characteristics from spatial variations by taking advantage of traditional registration and scene flow estimation as pretexts. A brief summary on the methods under this category is shown in Table \ref{tab:motion}.

\begin{table*}[]
    \centering
    \caption{Summary of motion-based point cloud SSL methods.}
    \label{tab:motion}
    \resizebox*{\linewidth}{!}{
    \begin{tabular}{|c|c|c|c|}\hline
        Year & Method & Sub-categories & Contributions  \\\hline
        2019 & PRNet \citep{wang2019prnet} & Registration & Pioneer for partial-to-partial point cloud registration enabling coarse-to-fine refinement \\
        2021 & Part mobility \citep{shi2021self} & Registration & Converting points to trajectories to derive the rigid transformation hypotheses \\
        2022 & SuperLine3D \citep{zhao2022superline3d} & Registration & Obtaining precise line representation under arbitrary scale perturbations \\
        2022 & DVD \citep{liu2022self} & Registration & Learning local and global point embedding jointly \\\hline
        2020 & PointPWC-Net \citep{wu2020pointpwc} & Scene flow estimation & Discretizing cost volume onto 3D point clouds in a coarse-to-fine fashion \\
        2020 & Just go with the flow \citep{mittal2020just} & Scene flow estimation & Optimizing two SSL losses based on nearest neighbors and cycle consistency \\
        2021 & Self-Point-Flow \citep{li2021self} & Scene flow estimation & Converting pseudo label matching problem as optimal transport task \\\hline

    \end{tabular}}

\end{table*}

\subsubsection{Registration}
\begin{figure}[htbp]
    \centering
    \includegraphics[width=0.97\linewidth]{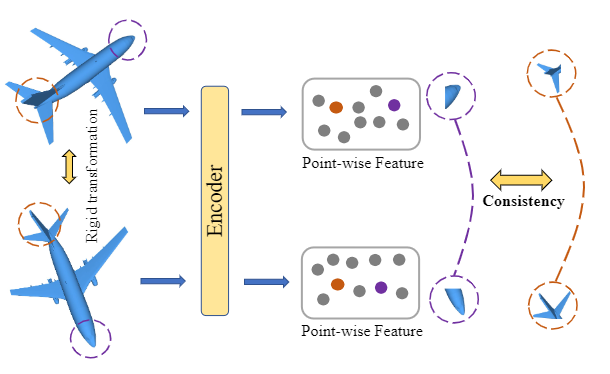}
    \caption{Demonstration of deep versatile descriptors. The input consists of two point clouds before and after rigid transformations, where the common point components are utilized to train the encoder for global and local representation learning. The figure is adapted from \citep{liu2022self}.}  
    \label{fig:DVD}
\end{figure}

Point cloud registration is the task to merge two point clouds $X$ and $Y$ into a globally consistent coordinate system via estimating the rigid transformation matrix, which can be formulated as:

\begin{equation}
    R, t=\underset{R \in S O(3), t \in \mathbb{R}^{3}}{\arg \min } \|\psi (R X + t) - \psi(Y) \|_{2} .
\end{equation}

where $R \in S O(3)$ and $t \in \mathbb{R}^{3}$ indicate rotation matrix and translation vector, respectively; $\psi$ is the feature extraction network learning the hierarchical informative features from dynamic point clouds. Unlike the classic ICP registration method \citep{besl1992method} which iteratively searches correspondences and estimates rigid transformation, SSL registration can obtain informative point cloud features without high-quality ground-truth correspondences.

PRNet \citep{wang2019prnet} is a partial-to-partial registration method that enables coarse-to-fine refinement iteratively. Based on co-contextual information, the framework boils down the registration problem as a key point detection task, which aims to recognize the matching points from two input clouds. Shi \citep{wang2019prnet} presented a part mobility segmentation approach to understand the essential attributes of the dynamic object. Instead of directly processing the sequential point clouds, the raw input is converted to trajectories by point correspondence between successive frames to derive rigid transformation hypotheses. Analogously, Zhao et al. \citep{zhao2022superline3d} proposed an SSL line segmentation and description for LiDAR point clouds, called SuperLine3D, providing applicable line features for global registration without any prior hints. Compared to point embedding constrained by limited resolution, this segmentation model is capable of obtaining precise line representation under arbitrary scale perturbations. 

Motivated by the observation that the local distinctive geometric structures of two subsets of point clouds can improve representations, Liu et al. \citep{liu2022self} introduced the deep versatile descriptors (DVDs) which learn local and global point embeddings jointly. As shown in Fig. \ref{fig:DVD}, the co-occurring point cloud local regions, which retain the structural knowledge under rigid transformations, are regarded as the input of DVD to extract latent geometric patterns restrained by local consistency loss. To further enhance the model's capability of transformation awareness, reconstruction and normal estimation are added as auxiliary tasks for better alignment.

\subsubsection{Scene flow estimation}

\begin{figure}[htbp]
    \centering
    \includegraphics[width=0.97\linewidth]{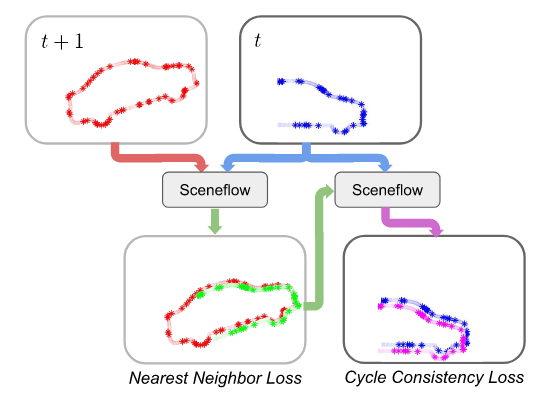}
    \caption{Demonstration of just going with the flow. The nearest neighbor loss is utilized to push the predicted flow (green) close to the pseudo-ground truth (red) of the frame at $t+1$. The cycle consistency loss is the penalty term to estimate the flow between predicted points (green) in the opposite direction to the original points (blue) in frame at $t$ for temporal alignment. The figure is adapted from \citep{mittal2020just}.}  
    \label{fig:just_go}
\end{figure}

Scene flow estimation is a vital computer vision task. For point clouds, its objective is to estimate the motion of objects by computing dense correspondences between consecutive LiDAR scans of a scene over time. The variation of points can be represented as 3D displacement vectors to describe the motions in terms of scene flow.

Wu et al. \citep{wu2020pointpwc} introduced the notion of cost volume and proposed a learnable point-based network called PointPWC-Net. The cost volume is discretized as input point pairs to reduce computational complexity; additionally, an efficient upsampling strategy and wrap layers are employed. Mittal et al. \citep{mittal2020just} proposed a novel SSL scene flow estimation network to achieve safe navigation during interactions with highly dynamic environments by optimizing two loss components based on the nearest neighbors and cycle consistency. As shown in Fig. \ref{fig:just_go}, the nearest neighbor loss encourages the points predicted based on current moment $t$ flowing toward occupied regions of the future frame at $t+1$. The cycle consistency loss ensures that the points of the future frame $t+1$ can be restored in the reverse direction back to frame $t$ to avoid degenerate solutions by maintaining temporal consistency. Self-Point-Flow \citep{li2021self} employs more than 3D point coordinates, surface normal, and color in one-to-one matching to generate pseudo labels and formulates the pseudo label generation issue as an optimal transport problem. It leverages a random walk module to refine annotation quality by imposing local alignment.

\section{Downstream tasks}\label{sec:downstream}
One of the primary objectives of SSL is to pre-train a backbone network and transfer it to solve the problems in downstream tasks. Therefore, performance of the model in downstream tasks could reflect the effectiveness of SSL to a certain degree. The evaluation criteria indicate whether the SSL methods can extract useful knowledge from pretext tasks with large-scale unlabeled point cloud data. In this section, we introduce four commonly used downstream tasks and provide the widely used evaluation metrics. In addition, we summarize and compare the performance of the aforementioned representative SSL methods in the corresponding downstream tasks.

\subsection{Object classification}
Object classification is a fundamental and prevalent downstream task that requires the model to output a most likely label for the given point cloud object to assess the overall semantic awareness of the pre-trained model. The two commonly used metrics for this task are Overall Accuracy (OA) and Mean Class Accuracy (mAcc). OA is the ratio of correctly classified objects to the total number of objects, and mAcc is the average of each class's accuracy. Object classification can be divided into three protocols based on task settings:

\begin{itemize}
    \item \textbf{Few-shot:} Few-shot learning (FSL) is a challenging task that involves training with limited information provided by the downstream dataset. Specifically, the $n$-way, $m$-shot setting is employed, where $n$ is the number of classes randomly selected from the dataset and $m$ is the number of objects randomly sampled for each class. The trained model is evaluated on the test split. Few-shot protocol performance of reviewed SSL methods is shown in Table \ref{tab:few-shot}.
    
    \item \textbf{Fine-tuning:} The pre-trained feature extractor serves as the initial downstream backbone encoder, and the entire network is re-trained in a supervised manner with labels from the downstream datasets. Fine-tuning protocol performance of proposed SSL methods is presented in Table \ref{tab:fine-tuning}.
    
    \item \textbf{Linear classification:} The pre-trained feature extractor is frozen by stopping the backpropagation gradients. Linear classifiers are trained in a supervised manner with downstream datasets. Linear classification protocol performance of proposed SSL methods is shown in Table \ref{tab:linear_classification}.
\end{itemize}

\begin{table}[]
    \centering
    \caption{Summary of few-shot protocol performance of representative SSL methods on ModelNet40 \citep{wu20153d} and ScanObjectNN \citep{uy2019revisiting}. The results are reported in terms of OA (\%).}
    \label{tab:few-shot}
    \resizebox*{\linewidth}{!}{
    \begin{tabular}{*{6}{c}}
        \toprule
        \multirow{2.5}*{Method} & \multirow{2.5}*{Backbone} & \multicolumn{2}{c}{5-way} & \multicolumn{2}{c}{10-way} \\ \cmidrule{3-4} \cmidrule{5-6} && 10-shot & 20-shot & 10-shot & 20-shot \\
        \midrule
        \multicolumn{6}{c}{ModelNet40} \\
        Point-MAE \citep{pang2022masked} & Transformer & 96.3 & 97.8 & 92.6 & 95.5 \\
        Point-BERT \citep{yu2021point} & Transformer & 94.6 & 96.3 & 91.0 & 92.7 \\    
        OcCo \citep{wang2021unsupervised} & PointNet & 89.7 & 92.4 & 89.3 & 89.7 \\
        OcCo \citep{wang2021unsupervised} & DGCNN & 90.6 & 92.5 & 82.9 & 86.5 \\
        OcCo \citep{wang2021unsupervised} & Transformer & 94.0 & 95.9 & 89.4 & 92.4 \\
        3D jigsaw \citep{sauder2019self} & PointNet & 66.5 & 69.2 & 56.9 & 66.5 \\
        3D jigsaw \citep{sauder2019self} & DGCNN & 34.3 & 42.2 & 26.0 & 29.9 \\
        MaskSurf \citep{zhang2022masked} & Transformer & 96.5 & 98.0 & 93.0 & 95.3 \\
        MaskPoint \citep{liu2022masked} & Transformer & 95.0 & 97.2 & 91.4 & 93.4 \\
        
        \midrule
        \multicolumn{6}{c}{ScanObjectNN} \\
        Point-MAE \citep{pang2022masked} & Transformer & 63.9 & 77.0 & 53.6 & 61.6 \\
        OcCo \citep{wang2021unsupervised} & PointNet & 70.4 & 72.2 & 54.8 & 61.8 \\
        OcCo \citep{wang2021unsupervised} & DGCNN & 72.4 & 77.2 & 57.0 & 61.6 \\ 
        3D jigsaw \citep{sauder2019self} & PointNet & 58.6 & 67.6 & 53.6 & 48.1 \\
        3D jigsaw \citep{sauder2019self} & DGCNN & 65.2 & 72.2 & 45.6 & 48.2 \\
        MaskSurf \citep{zhang2022masked} & Transformer & 65.3 & 77.4 & 53.8 & 63.2 \\
        
        \bottomrule
    \end{tabular}}
\end{table}

\begin{table*}[]
    \centering
    \caption{Summary of fine-tuning protocol performance of representative SSL methods on ModelNet40 \citep{wu20153d} and ScanObjectNN \citep{uy2019revisiting}. ScanObjectNN has three challenges. The results are reported in terms of OA (\%).}
    \label{tab:fine-tuning}
    \resizebox*{\linewidth}{!}{
    \begin{tabular}{*{9}{c}}
        \toprule
        \multirow{2.5}*{Method} & \multirow{2.5}*{Year} & \multirow{2.5}*{Pretext type} & \multirow{2.5}*{Backbone} & \multirow{2.5}*{Pre-train dataset} & \multirow{2.5}*{ModelNet40} & \multicolumn{3}{c}{ScanObjectNN} \\ \cmidrule{7-9} &&&&&& OBJ-BG & OBJ-ONLY & PB-T50-RS \\
        \midrule
        Supervised & \makecell[c]{2017 \\ 2017 \\ 2019 \\ 2017} & - & \makecell[c]{PointNet \citep{qi2017pointnet} \\ PointNet++ \citep{qi2017pointnet++} \\ DGCNN \citep{wang2019dynamic} \\ Transformer \citep{vaswani2017attention}} & - & \makecell[c]{89.2 \\ 90.7 \\ 92.9 \\ 91.4} & \makecell[c]{73.3 \\ 82.3 \\ 82.8 \\ 79.86} & \makecell[c]{79.2 \\ 84.3 \\ 86.2 \\ 80.55} & \makecell[c]{68.0 \\ 77.9 \\ 78.1 \\ 77.24} \\
        \midrule
        
        Point-MAE \citep{pang2022masked} & 2022 & Reconstruction & Transformer & ShapeNet & 93.8 & 90.02 & 88.29 & 85.18 \\ 
        Point-BERT \citep{yu2021point} & 2021 & Reconstruction & Transformer & ShapeNet & 93.2 & 87.43 & 88.12 & 83.07 \\ 
        3D jigsaw \citep{sauder2019self} & 2019 & Reconstruction & DGCNN & ShapeNet & 92.4 & 82.0 & 82.1 & - \\
        MaskSurf \citep{zhang2022masked} & 2022 & Reconstruction & Transformer & ShapeNet & 93.40 & 91.22 & 89.17 & 85.81 \\ 
        CloudContext \citep{sauder2019context} & 2019 & Reconstruction & DGCNN & ShapeNet & 90.8 & - & - & - \\
        UAE \citep{zhang2022upsampling} & 2022 & Reconstruction & DGCNN & ShapeNet & 93.2 & - & - & - \\
        MD \citep{sun2022self} & 2022 & Reconstruction & DGCNN & ModelNet40 & 93.39 & - & - & - \\
        Self-correction \citep{chen2021shape} & 2021 & Reconstruction & PointNet & ShapeNet & 90.0 & - & - & - \\
        Self-correction \citep{chen2021shape} & 2021 & Reconstruction & RSCNN & ShapeNet & 93.0 & - & - & - \\
        MaskPoint \citep{liu2022masked} & 2022 & Reconstruction & Transformer & ShapeNet & 93.8 & 88.1 & 89.3 & 84.3 \\
        \midrule
        Info3D \citep{sanghi2020info3d} & 2020 & Contrast & PointNet & ShapeNet & 90.20 & - & - & - \\
        Info3D \citep{sanghi2020info3d} & 2020 & Contrast & DGCNN & ShapeNet & 93.03 & - & - & - \\
        \midrule
        OcCo \citep{wang2021unsupervised} & 2021 & Alignment & PointNet & ModelNet40 & 90.1 & - & - & - \\
        OcCo \citep{wang2021unsupervised} & 2021 & Alignment & DGCNN & ModelNet40 & 93.0 & 82.1 & 83.2 & - \\
        Cross-view \citep{jing2021self} & 2021 & Alignment & DGCNN & ModelNet40 & 93.0 & 82.2 & 83.0 & - \\
        Multi-view rendering \citep{tran2022self} & 2022 & Alignment & PointNet & ModelNet40 & 89.5 & 78.5 & 80.5 & - \\
        Multi-view rendering \citep{tran2022self} & 2022 & Alignment & DGCNN & ModelNet40 & 93.2 & 84.5 & 84.3 & - \\
        STRL \citep{huang2021spatio} & 2021 & Alignment & DGCNN & ShapeNet & 93.1 & - & - & - \\
        \bottomrule
    \end{tabular}}
\end{table*}

\begin{table}[]
    \centering
    \caption{Summary of linear classification protocol performance of representative SSL methods on ModelNet10/40 \citep{wu20153d}. The results are reported in terms of OA (\%).}
    \label{tab:linear_classification}
    \resizebox*{\linewidth}{!}{
    \begin{tabular}{cccc}
    \toprule
    Method & Pretext type & Backbone & ModelNet10/40 \\ 
    \midrule
    Point-MAE \citep{pang2022masked} & Reconstruction & Transformer & - / 91.41 \\ 
    Orientation estimation \citep{poursaeed2020self} & Reconstruction & PointNet & - / 88.6 \\
    Orientation estimation \citep{poursaeed2020self} & Reconstruction & DGCNN & - / 90.75 \\
    3D jigsaw \citep{sauder2019self} & Reconstruction & PointNet & 91.61 / 87.31 \\    
    3D jigsaw \citep{sauder2019self} & Reconstruction & DGCNN & 94.52 / 90.64 \\
    MaskSurf \citep{zhang2022masked} & Reconstruction & Transformer & - / 92.26 \\ 
    CloudContext \citep{sauder2019context} & Reconstruction & DGCNN & 94.5 / 89.3 \\
    UAE \citep{zhang2022upsampling} & Reconstruction & DGCNN & 95.6 / 92.9 \\
    Pose Disentanglement \citep{tsai2022self} & Reconstruction & PointNet & - / 90.1 \\
    Pose Disentanglement \citep{tsai2022self} & Reconstruction & DGCNN & - / 92.0 \\
    CP-Net \citep{xu2022cp} & Reconstruction & RSCNN & - / 91.9 \\
    FoldingNet \citep{yang2018foldingnet} & Reconstruction & GNN & 94.4 / 88.4 \\
    Self-correction \citep{chen2021shape} & Reconstruction & PointNet & 93.3 / 89.9 \\
    Self-correction \citep{chen2021shape} & Reconstruction & RSCNN & 95.0 / 92.4 \\ 
    PC-GAN \citep{li2018point} & Reconstruction & GAN & - / 87.5 \\
    \midrule
    Info3D \citep{sanghi2020info3d} & Contrast & PointNet & - / 89.8 \\
    Info3D \citep{sanghi2020info3d} & Contrast & DGCNN & - / 91.6 \\
    AFSRL \citep{lu2022joint} & Contrast & GNN & - / 91.5 \\
    Contrasting and clustering \citep{zhang2019unsupervised} & Contrast & DGCNN & 93.8 / 86.8 \\
    Hard negatives \citep{du2021self} & Contrast & DGCNN & - / 89.6 \\
    \midrule
    OcCo \citep{wang2021unsupervised} & Alignment & DGCNN & - / 89.2 \\
    Cross-view \citep{jing2021self} & Alignment & GNN & - / 89.8 \\
    Multi-view rendering \citep{tran2022self} & Alignment & PointNet & - / 89.7 \\
    Multi-view rendering \citep{tran2022self} & Alignment & DGCNN & - / 91.7 \\
    STRL \citep{huang2021spatio} & Alignment & PointNet & - / 88.3 \\
    STRL \citep{huang2021spatio} & Alignment & DGCNN & - / 90.9 \\
    \midrule
    PRNet \citep{wang2019prnet} & Motion & DGCNN & - / 85.2 \\
    \bottomrule
    \end{tabular}}

\end{table}

\subsection{Part segmentation}
Part segmentation is a fine-grained task that aims to distinguish and separate various components of an object, such as plane wings or desk legs. This task usually requires a model that can extract local point-level features more effectively than the overall discriminative ability required for object recognition. The popular evaluation criteria of point cloud part segmentation is the mean Intersection over Union ($\rm mIoU$), which computes the ratio of the intersection of the predicted and ground truth part labels to the union of the two, across all categories ($\rm mIoU_{C}$) or all instances ($\rm mIoU_{I}$). Table \ref{tab:part_segmentation} summarizes the results of part segmentation on the ShapeNetPart dataset based on SSL pre-training models and supervised fine-tuning in terms of $\rm mIoU_{C}$ (\%), $\rm mIoU_{I}$ (\%).

\begin{table}[]
    \centering
    \caption{Summary of performance of representative methods on part segmentation using ShapeNetPart \citep{armeni20163d}.}
    \label{tab:part_segmentation}
    \resizebox*{\linewidth}{!}{
    \begin{tabular}{*{5}{c}}
        Method & Type & Backbone & $\rm mIoU_{C}$ & $\rm mIoU_{I}$  \\
        \toprule
        Supervised & - & \makecell[c]{PointNet \citep{qi2017pointnet} \\ PointNet++ \citep{qi2017pointnet++} \\ DGCNN \citep{wang2019dynamic} \\ Transformer \citep{vaswani2017attention}} & \makecell[c]{83.39 \\ 81.85 \\ 82.33 \\ 83.42} & \makecell[c]{83.7 \\ 85.1 \\ 85.2 \\ 85.1} \\
        \midrule
        Point-MAE \citep{pang2022masked} & Reconstruction & Transformer & 84.19 & 86.1 \\
        Point-BERT \citep{yu2021point} & Reconstruction & Transformer & 84.11 & 85.6 \\
        3D jigsaw \citep{sauder2019self} & Reconstruction & PointNet & - & 82.2 \\
        3D jigsaw \citep{sauder2019self} & Reconstruction & DGCNN & - & 85.3 \\
        MaskSurf \citep{zhang2022masked} & Reconstruction & Transformer & 84.36 & 86.1 \\
        CloudContext \citep{sauder2019context} & Reconstruction & DGCNN & - & 81.5 \\
        UAE \citep{zhang2022upsampling} & Reconstruction & DGCNN & - & 85.6 \\
        Pose Disentanglement \citep{tsai2022self} & Reconstruction & PointNet & - /83.8 \\
        Pose Disentanglement \citep{tsai2022self} & Reconstruction & DGCNN & - / 85.1 \\
        MD \citep{sun2022self} & Reconstruction & DGCNN & - & 85.5 \\
        Self-correction \citep{chen2021shape} & Reconstruction & PointNet & - & 84.1 \\
        Self-correction \citep{chen2021shape} & Reconstruction & RSCNN & - & 85.2 \\
        MaskPoint \citep{liu2022masked} & Reconstruction & Transformer & 84.4 & 86.0 \\
        \midrule
        AFSRL \citep{lu2022joint} & Contrast & GNN & - / 85.7 \\
        Hard negatives \citep{du2021self} & Contrast & DGCNN & - / 82.3 \\
        PointContrast \citep{xie2020pointcontrast} & Contrast & U-Net & - & 85.1 \\
        \midrule
        OcCo \citep{wang2021unsupervised} & Alignment & PointNet & - & 83.4 \\
        OcCo \citep{wang2021unsupervised} & Alignment & DGCNN & - & 85.0 \\
        Cross-view \citep{jing2021self} & Alignment & DGCNN & 79.1 & 83.7 \\
        Multi-view rendering \citep{tran2022self} & Alignment & PointNet & - & 83.3 \\
        Multi-view rendering \citep{tran2022self} & Alignment & DGCNN & - & 84.7 \\
        \midrule
        PRNet \citep{wang2019prnet} & Motion & DGCNN & 78.8 / 82.5 \\
        \bottomrule
    \end{tabular}}
\end{table}

\subsection{Semantic segmentation}
Semantic segmentation requires a model to assign a semantic label to each points in the point cloud in order to group meaningful regions. It is frequently implemented on complicated outdoor or indoor scenes with background noise. $\rm mIoU$, OA, and mAcc are commonly employed as estimation indicators to judge the feature extraction capability of pre-training models on the S3DIS dataset \citep{armeni20163d}, which contains six large-scale indoor venues, with the following two protocols. Performance of representative methods on semantic segmentation under the two protocols are shown in Table \ref{tab:semantic_segmentation}.

\begin{itemize}
    \item \textbf{Area 5 test}: The SSL pre-trained model is fine-tuned on all areas except the largest area 5, which is chosen as the test set.
    \item \textbf{Six-fold cross validation}: Areas 1-6 are selected in turn as the test set and fine-tuned in the remaining 5 areas.
\end{itemize}

\begin{table}[]
    \centering
    \caption{Summary of performance of representative methods on semantic segmentation using S3DIS \citep{armeni20163d}.}
    \label{tab:semantic_segmentation}
    \resizebox*{\linewidth}{!}{
    \begin{tabular}{*{6}{c}}
        Method & Type & Backbone & OA & mAcc & mIoU  \\
        \toprule
        Supervised & - & \makecell[c]{PointNet \citep{qi2017pointnet} \\ DGCNN \citep{wang2019dynamic} \\ Transformer \citep{vaswani2017attention}} & \makecell[c]{78.6 \\ 84.1 \\ 86.8} & \makecell[c]{49.0 \\ - \\ 68.6} & \makecell[c]{47.7 \\ 56.1 \\ 60.0} \\
        \midrule
        \multicolumn{6}{c}{Area 5 test} \\
        Point-MAE \citep{pang2022masked} & Reconstruction & Transformer & 87.4 & 69.4 & 61.0 \\
        OcCo \citep{wang2021unsupervised} & Alignment & PointNet & - & 83.6 & 44.5 \\
        OcCo \citep{wang2021unsupervised} & Alignment & DGCNN & - & 87.0 & 49.5 \\
        3D jigsaw \citep{sauder2019self} & Reconstruction & PointNet & - & 82.5 & 43.6 \\
        3D jigsaw \citep{sauder2019self} & Reconstruction & DGCNN & - & 86.8 & 48.2 \\
        MaskSurf \citep{zhang2022masked} & Reconstruction & Transformer & 88.3 & 69.9 & 61.6 \\
        PointContrast \citep{xie2020pointcontrast} & Contrast & SR-UNet & - & 77.0 & 70.9 \\
        Contrastive Scene Contexts \citep{hou2021exploring} & Contrast & DGCNN & - & - & 73.8 \\
        DepthConstrast \citep{zhang2021self} & Contrast & PointNet++ & - & 72.1 & 64.8 \\
        Multi-view rendering \citep{tran2022self} & Alignment & PointNet & - & 85.0 & 46.7 \\
        Multi-view rendering \citep{tran2022self} & Alignment & DGCNN & - & 87.0 & 49.9 \\
        
        \midrule
        \multicolumn{6}{c}{Six-fold cross validation} \\
        OcCo \citep{wang2021unsupervised} & Alignment & PointNet & 82.0 & - & 54.9 \\
        OcCo \citep{wang2021unsupervised} & Alignment & DGCNN & 84.6 & - & 58.0 \\
        3D jigsaw \citep{sauder2019self} & Reconstruction & PointNet & 80.1 & - & 52.6 \\
        3D jigsaw \citep{sauder2019self} & Reconstruction & DGCNN & 84.1 & - & 55.6 \\
        CloudContext \citep{sauder2019context} & Reconstruction & DGCNN & 78.9 & - & 47.6 \\
        Multi-view rendering \citep{tran2022self} & Alignment & PointNet & - & 83.2 & 52.1 \\
        Multi-view rendering \citep{tran2022self} & Alignment & DGCNN & - & 87.5 & 59.0 \\
        
        \bottomrule
    \end{tabular}}
\end{table}

\subsection{Object detection}
Object detection is a task that involves localizing the 6 Degrees-of-Freedom (DoF) bounding box of an object and differentiating its category in a complex scene. The evaluation metric used is the average precision (AP), which measures the precision of the 3D bounding box at various recall levels. The threshold is usually set to 0.25 and 0.5. Table \ref{tab:object_detection} summarizes the object detection performance of the SSL pre-training models on the SUN RGB-D \citep{song2015sun} and ScanNet \citep{dai2017scannet} datasets.

\begin{table}[]
    \centering
    \caption{Summary of performance of representative methods on object detection using SUN RGB-D \citep{song2015sun} and ScanNet \citep{dai2017scannet}. The pre-training input only contains the point cloud geometry.}
    \label{tab:object_detection}
    \resizebox*{\linewidth}{!}{
    \begin{tabular}{*{7}{c}}
        \multirow{2.5}*{Method} & \multirow{2.5}*{Type} & \multirow{2.5}*{Backbone} & \multicolumn{2}{c}{SUN RGB-D} & \multicolumn{2}{c}{ScanNet} \\ \cmidrule{4-5} \cmidrule{6-7} &&& $\rm AP_{25}$ & $\rm AP_{50}$ & $\rm AP_{25}$ & $\rm AP_{50}$ \\
        \toprule
        Point-BERT \citep{yu2021point} & Reconstruction & Transformer & - & - & 61.0 & 38.3 \\
        MaskPoint \citep{liu2022masked} & Reconstruction & Transformer & - & - & 64.2 & 42.0 \\
        PointContrast \citep{xie2020pointcontrast} & Contrast & SR-UNet & 57.5 & 34.8 & 59.2 & 38.0 \\
        PointContrast \citep{xie2020pointcontrast} & Contrast & VoteNet & 59.2 & 38.0 & 57.5 & 34.8 \\
        DepthConst \citep{zhang2021self} & Contrast & PointNet++ & - & - & 61.3 & - \\
        DepthConst \citep{zhang2021self} & Contrast & VoteNet & 64.0 & 42.9 & 61.6 & 35.5 \\
        DepthConst \citep{zhang2021self} & Contrast & H3DNet & 69.0 & 50.0 & 63.5 & 43.4 \\
        Multi-view rendering \citep{tran2022self} & Alignment & DGCNN & 58.1 & 35.1 & 60.3 & 39.2 \\
        STRL \citep{huang2021spatio} & Alignment & VoteNet & 58.2 & - & - & - \\
        \bottomrule
    \end{tabular}}
\end{table}

\section{Future directions}\label{sec:future}
Although self-supervised learning has shown great success for point cloud processing, we have identified some of its deficiencies and limitations. We argue that SSL should not be studied in isolation but rather in conjunction with advanced techniques from other domains. In this section, we discuss a number of future research directions that have the potential to improve the SSL learning capability and performance on downstream tasks.

\subsection{Few-shot and zero-shot learning}
There have been a good number of publicly available, labelled datasets for SSL research. However, real scenarios often face the data shortage or quality challenges, such as damaged labels, missing information, and uneven assortment. Few-shot learning (FSL) \citep{garcia2017few} is considered as a potential solution that allows the network to train under the situations with very small amount of data. It is also possible to identify new sample types that have not been seen before in a test task without training samples. This method is often referred to as the zero-shot learning (ZSL). Both SSL and FSL (ZSL) \citep{romera2015embarrassingly} can free models from the reliance on large annotated datasets and reduce the cost. In addition, the combination of these two could potentially improve the generalization capability the models.



\subsection{Multiple modality interaction and fusion}
Despite of the assorted modalities in many existing datasets, for example, for outdoor autonomous driving \citep{geiger2012we,sun2020scalability,caesar2020nuscenes}, researchers normally only focus on and make use of the point cloud data while ignoring the connections and alignment relationships with data of other modalities. We have seen some recent research works design models \citep{vora2020pointpainting,wang2021pointaugmenting,li2022deepfusion} for multi-model data alignment and fusion, primarily point clouds and images. We anticipate more research to focus on cross-modal SSL with more diverse modalities, e.g., natural language, radar and voice, exploiting the unique characteristics of each modality and the synergy among them to build transportation systems, e.g. autonomous driving and traffic scene analysis, with more artificial general intelligence.

\subsection{Hierarchical feature extraction}
To cope with sophisticated downstream tasks with somehow conflicting objectives, for example, object classification which requires overall semantic understanding and part segmentation which requires fine-grained geometrical awareness, SSL models should have the capability for both global perception and local analysis. This necessitates hierarchical feature extraction; in particular, interactions between feature representations on different levels in the hierarchy need to be considered to discover the implicit relations. Therefore, we suggest that hierarchical feature extraction should be embedded in the SSL paradigm to improve the model's capability to capture both global and local features from point clouds.

\subsection{Multiple tasks pre-training}
Up to now, most point cloud SSL methods have only one specific pre-training pretext while few works train diverse tasks concurrently. The main resistance is that multi-tasking has to consider the compatibility and synergy between various pretexts simultaneously, and fit each loss item for steady parameter updating. This is also one of the very reasons why a model performs well on one downstream task but not others. Indeed, distinct proxies could provide useful information from various perspectives of point clouds so that jointly training multiple tasks could facilitate the network to learn more comprehensive representations; obviously, more research on multi-task SSL is needed to push the research one step further.

\subsection{Theory and interpretability}
Similar to traditional deep learning, point cloud SSL lacks sufficient theoretical support and has poor interpretation. The process of model training is conducted as a black-box, making it difficult for human users to analyze the results. Most of the technical works demonstrate their contributions via ablation studies and draw conclusions empirically. Such 'tried and tested' methods do not have theoretical support and are therefore difficult to verify, generalize and replicate. We suggest that future studies should include more inquiries into explainable theory, for example, the well-established theories from mutual information \citep{sayed2018cross} or causal inference \citep{wang2022causal}, which can be applied in the design of network structures and loss functions.

\section{Conclusion} \label{sec:conclusion}
Point cloud self-supervised learning fundamentally moves away from models' dependency on manual annotations. The learning paradigm focuses on the design of pre-training pretext tasks to enable the models to extract effective features and achieves performance competitive to the supervised learning paradigms in many downstream tasks. This paper extensively surveys recent representative deep neural network-based methods for self-supervised learning from point cloud data. A novel taxonomy is proposed to systematically classify the current research, especially the works publishes in the recent three years. Besides detailed analysis on the representative methods, we provide summaries on the commonly used datasets and performance comparison to make the survey more comprehensive. Future research directions are also discussed to hopefully provide an insightful view on the issues that the research community should pay attention to. We hope that our work provides a valuable reference on point cloud SSL research and could motivate researchers to further explore this promising topic.

\section{Acknowledgement}
This work received financial support from Jiangsu Industrial Technology Research Institute (JITRI) and Wuxi National Hi-Tech District (WND).

\bibliographystyle{cas-model2-names}
\bibliography{New_SSL_Survey.bib}
\end{document}